\documentclass[journal]{IEEEtran}
\usepackage{tabto}
\usepackage{algpseudocode}
\usepackage{algorithm}
\usepackage{paralist}
\usepackage{cite}
\usepackage{hyperref}
\usepackage{amsmath,amssymb,amsfonts}
\usepackage{subcaption}
\usepackage{tikz}
\usepackage{array}
\usetikzlibrary{positioning, shapes.geometric, arrows}
\usetikzlibrary{shapes,arrows}
\usetikzlibrary{positioning, arrows.meta}
\usepackage{graphicx}
\usepackage{comment}

\usepackage{textcomp}
\usepackage{xcolor}
\def\BibTeX{{\rm B\kern-.05em{\sc i\kern-.025em b}\kern-.08em
    T\kern-.1667em\lower.7ex\hbox{E}\kern-.125emX}}

\newcommand{\nix}[1]{}

\tikzstyle{block_yellow} = [rectangle, draw, fill=yellow!30,
    text width=10em, text centered, rounded corners, minimum height=3em]
\tikzstyle{block_green} = [rectangle, draw, fill=green!30,
    text width=10em, text centered, rounded corners, minimum height=3em]
\tikzstyle{block_purple} = [rectangle, draw, fill=purple!30,
    text width=10em, text centered, rounded corners, minimum height=3em]
\tikzstyle{block_blue} = [rectangle, draw, fill=blue!30,
    text width=10em, text centered, rounded corners, minimum height=3em]
\tikzstyle{block_pink} = [rectangle, draw, fill=pink!30,
    text width=10em, text centered, rounded corners, minimum height=3em]
\tikzstyle{block_grey} = [rectangle, draw, fill=gray!30,
    text width=10em, text centered, rounded corners, minimum height=3em]

\tikzstyle{arrow} = [thick,->,>=stealth]

\begin{document}
\title{Revolutionizing Communication with Deep Learning and XAI for Enhanced Arabic Sign Language Recognition}

\author{\IEEEauthorblockN{Mazen Balat$^\dag$,~~~ Rewaa Awaad$^\dag$,~~~ Ahmed B. Zaky$^\dag$,~~~Salah A. Aly$^\S$ \\ }
\IEEEauthorblockA{
\small{$^\dag$CS \& IT Department, Egypt-Japanese University of Science \& Technology, Alexandria,  Egypt\\
$^\S$Faculty of Computing and Data Science, Badya University, Giza, Egypt\\
$^\S$Computer Science Section, Faculty of Science, Fayoum University, Fayoum, Egypt \\
}}}

\maketitle

\begin{abstract}
This study introduces an integrated approach to recognizing Arabic Sign Language (ArSL) using state-of-the-art deep learning models such as MobileNetV3, ResNet50, and EfficientNet-B2. These models are further enhanced by explainable AI (XAI) techniques to boost interpretability. The ArSL2018 and RGB Arabic Alphabets Sign Language (AASL) datasets are employed, with EfficientNet-B2 achieving peak accuracies of 99.48\% and 98.99\%, respectively. Key innovations include sophisticated data augmentation methods to mitigate class imbalance, implementation of stratified 5-fold cross-validation for better generalization, and the use of Grad-CAM for clear model decision transparency. The proposed system not only sets new benchmarks in recognition accuracy but also emphasizes interpretability, making it suitable for applications in healthcare, education, and inclusive communication technologies.

\end{abstract}


\pagenumbering{arabic}

%\tableofcontents

\section{Introduction}

AI and machine learning are reshaping everyday life, sparking innovation across fields like healthcare, education, and social services \cite{lipizzi2024societal}. These technologies are expanding human potential, tackling social issues, and promoting inclusivity by helping to bridge communication barriers among diverse groups \cite{bovzic2023artifical}.

Sign language plays a crucial role as a communication method for individuals who are deaf or hard of hearing, enabling effective interaction within their communities and broader society \cite{othman2024sign}. Arabic Sign Language (ArSL), widely used in Arabic-speaking regions, is marked by unique gestures and expressions that capture the cultural richness and diversity of the Arab world \cite{hassan2024enhancing}. Developing precise and efficient ArSL recognition systems is vital for enhancing communication accessibility, thus promoting inclusivity and equal opportunities for people with hearing impairments.

The incorporation of automatic sign language recognition systems into daily life holds transformative potential across various sectors. In education, such systems facilitate real-time translation of educational materials, supporting deaf students in better engaging with their peers and educators \cite{almubayei2024sign}. Similarly, in healthcare, these systems improve communication between medical professionals and patients who use sign language, ensuring that critical information is effectively conveyed \cite{abdul2024empowering}. The deployment of these technologies in public spaces and consumer devices also raises awareness of sign language, fostering a more inclusive environment that encourages social interaction and reduces communication barriers \cite{yeratziotis2023making}.

Despite these benefits, developing ArSL recognition systems involves several challenges. The complexity of hand gestures, diverse signing styles, and the influence of environmental factors like lighting and background conditions can hinder recognition accuracy \cite{ibrahim2020advances}. Traditional approaches, such as using data gloves or relying on human interpreters, face limitations regarding practicality and scalability \cite{amin2023assistive}. Thus, there is an urgent need for advanced, automated solutions capable of addressing these challenges to ensure reliable and real-time recognition.

\begin{figure}[h]
\centerline{\includegraphics[width=8cm, height=5.5cm]{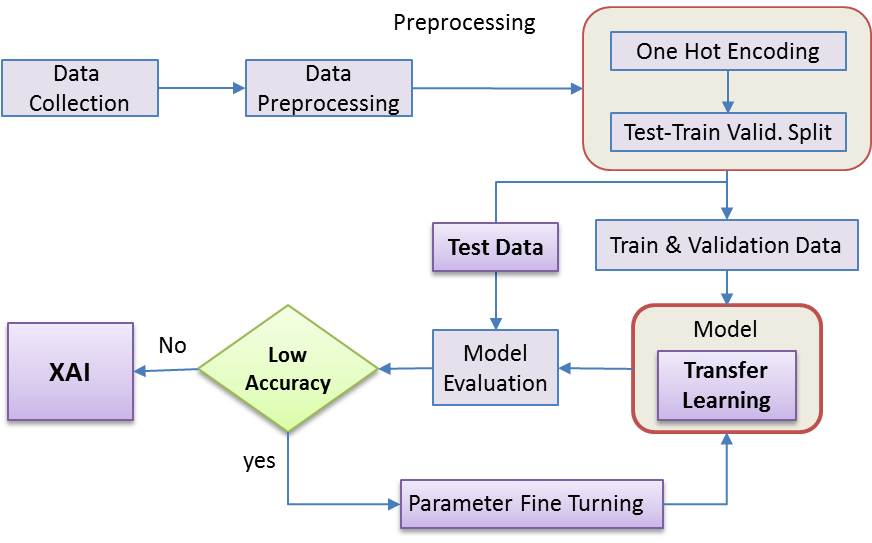}}
\caption{Work Flow Diagram}
\label{Work Flow Diagram}
\end{figure}

Recent advancements in deep learning have created new opportunities for sign language recognition. Convolutional Neural Networks (CNNs) and transformer-based models have shown effectiveness in processing complex visual and sequential data, making them well-suited for interpreting the intricate gestures of sign languages. CNNs, for instance, have demonstrated proficiency in analyzing images and videos of sign language, extracting critical features that aid in accurate classification \cite{buttar2023deep}. Additionally, transfer learning has been employed to enhance recognition performance even with limited datasets, leveraging pre-trained networks \cite{sharma2023continuous}.

A significant contribution of this research is the integration of explainable AI (XAI). By incorporating interpretability methods alongside recognition models, the study aims to achieve high accuracy while ensuring transparency in decision-making. Such transparency is particularly important in sensitive applications like healthcare and education, where understanding the rationale behind model predictions is essential. Moreover, XAI is reshaping everyday life by making AI systems more accessible and comprehensible to the general public. This helps users understand AI's decision-making processes, promoting greater trust and acceptance. In sectors such as finance, law enforcement, and customer service, XAI plays a critical role in ensuring that decisions are fair, unbiased, and accountable. As AI becomes more integrated into daily life, its ability to explain decisions will be pivotal for ethical considerations and societal acceptance \cite{dwivedi2023explainable}.

As depicted in Fig.~\ref{Work Flow Diagram}, the proposed pipeline comprises several stages, from data collection and preprocessing to model training, evaluation, and the integration of explainable AI techniques.

%\begin{figure*}[t]
%\centerline{\includegraphics[width=6in, height=2.5in]{fig/workflow3rotate.jpg}}
%\caption{Work Flow Diagram}
%\label{Work Flow Diagram}
%\end{figure*}

The main contributions of this research can be summarized as follows:

\begin{enumerate}
    \item The proposed system demonstrates superior recognition accuracy compared to existing state-of-the-art models, proving its efficacy in recognizing Arabic sign language gestures.
    \item Explainable AI techniques have been incorporated to ensure transparency in model decisions, which is critical in fields such as healthcare and education where understanding prediction rationale is essential.
    \item The recognition framework is adaptable to other sign languages, extending its applicability beyond ArSL. This adaptability enables usage across different languages and gestures.
    \item Through comprehensive preprocessing and model fine-tuning, the system maintains real-time recognition accuracy under diverse conditions, making it viable for real-world deployment.
\end{enumerate}

The structure of this paper is as follows: Section~\ref{sec:relatedworks} provides a comprehensive review of related literature on Arabic Sign Language (ArSL) recognition. Section~\ref{sec:dataset} introduces the datasets used in this study, specifically the ArASL2018 and RGB Arabic Alphabets Sign Language datasets. Section~\ref{sec:methodology} describes the methodology, including preprocessing steps and model architectures. Section~\ref{sec:model} outlines the model training process and detailed descriptions. Section~\ref{sec:evaluation} covers the evaluation metrics applied to assess model performance. Section~\ref{sec:XAI} delves into explainable AI techniques used to ensure transparent decision-making. Sections~\ref{sec:resultsMobileNet3} and ~\ref{sec:comparison} presents experimental results, including performance comparisons with state-of-the-art methods. Lastly, Section~\ref{sec:conclusion} concludes the study, summarizing the findings and suggesting future research directions, building upon our previous work in sign language research~\cite{Mazen2024sign}.

\section{Related Works}\label{sec:relatedworks}

Sign language communication tools, such as human interpreters, written communication methods, and Automatic Speech Recognition (ASR) systems, have provided essential support. However, these tools often lack comprehensive capabilities. The intricate and dynamic nature of sign languages—particularly Arabic Sign Language (ArSL)—poses significant challenges for standard machine learning models, which often struggle with the nuanced gestures and varied signing styles that characterize ArSL \cite{Balaha2023, app13010453, Shashidhar2023}.

In recent developments, deep learning has shown considerable promise for ArSL recognition, notably through transfer learning approaches. For instance, Hu et al. \cite{hu2022sign} developed a model using the ArSL2018 dataset, focusing on Arabic alphabet signs. They resized the images to 32x32 pixels and applied data augmentation, achieving 95\% accuracy using EfficientNetB4. Nonetheless, the model encountered issues related to class imbalances.

Al Ahmadi et al. \cite{al2024enhancing} utilized Convolutional Neural Networks (CNN) with transfer learning across three datasets (ASL-DS-I, ASL-DS-II, and ASL-DS-III), obtaining accuracy rates of 96.25\%, 95.85\%, and 97.02\%, respectively. Their work underscored CNNs' robustness in processing Arabic sign language data.

El Baz et al. \cite{el2024deep} focused on Arabic alphabet sign recognition using the RGB Arabic Alphabets Sign Language (AASL) dataset, collected from over 200 participants. After preprocessing, including background removal and data augmentation, they reported 99.4\% training accuracy and 97.4\% validation accuracy over 250 epochs.

\begin{table}[t]
\centering
\caption{Summary of Related Works on Arabic Sign Language Recognition}
\label{tab:relatedworks}
\begin{tabular}{|p{1.5cm}|p{1.5cm}|p{2.3cm}|p{1.4cm}|}
\hline
\textbf{Authors}         & \textbf{Dataset(s)}        & \textbf{Methodology}                & \textbf{Results}                            \\ \hline
Hu et al. \cite{hu2022sign}         & ArSL2018               & EfficientNetB4 with transfer learning & 95\% accuracy; faced class imbalance        \\ \hline
Al Ahmadi et al. \cite{al2024enhancing}  & ASL-DS-I, II, III      & CNN with transfer learning           & 96.25\%, 95.85\%, 97.02\% accuracies        \\ \hline
El Baz et al. \cite{el2024deep}     & AASL                   & CNN with data augmentation           & 99.4\% training, 97.4\% validation accuracy \\ \hline
Abdelghfar et al. \cite{abdelghfar2023model} & ArSL2018 (QSL subset) & QSLRS-CNN, resampling techniques     & 97.31\% accuracy                            \\ \hline
Al Nabih et al. \cite{alnabih2024arabic}   & ArSL2018               & Vision Transformers (ViT)            & 99.3\% accuracy                             \\ \hline
Lahiani et al. \cite{lahiani2024exploring} & ArSL2018               & InceptionV3, VGG16, MobileNetV2      & MobileNetV2 achieved 96\% accuracy          \\ \hline
Renjith et al. \cite{renjith2024sign}      & CSL, ArSL              & Spatio-temporal approach             & 90.87\% (CSL), 89.46\% (ArSL) accuracies    \\ \hline
Hassan et al. \cite{hassan2024detection}   & ArSL2018               & PCA, LDA, KNN                        & 86.4\% accuracy                             \\ \hline
\end{tabular}
\end{table}

Abdelghfar et al. \cite{abdelghfar2023model} explored Qur'anic Sign Language (QSL) recognition, using a subset of the ArSL2018 dataset. By employing Random Oversampling, SMOTE, and Random Undersampling to address class imbalances, their QSLRS-CNN model achieved a 97.31\% accuracy after 200 epochs.

Al Nabih et al. \cite{alnabih2024arabic} introduced Vision Transformers (ViT) for ArSL recognition, fine-tuning a pre-trained ViT model on the ArSL2018 dataset to reach 99.3\% accuracy. This demonstrated the potential of transformers to capture complex ArSL features, surpassing traditional CNNs.

Lahiani et al. \cite{lahiani2024exploring} evaluated various pre-trained CNN models, including InceptionV3, VGG16, and MobileNetV2, on the ArSL2018 dataset. MobileNetV2, enhanced with transfer learning, achieved the highest accuracy of 96\%.

Renjith et al. \cite{renjith2024sign} adopted a spatio-temporal approach that integrated both spatial and temporal features to capture sign language motions. Their method, tested on Chinese Sign Language (CSL) and ArSL, achieved accuracies of 90.87\% and 89.46\%, respectively, highlighting its effectiveness in handling dynamic sign language data.

Hassan et al. \cite{hassan2024detection} used traditional machine learning techniques to recognize ArSL on the ArSL2018 dataset. They applied greyscale conversion and feature extraction using PCA and LDA, with the K-Nearest Neighbors (KNN) classifier achieving the best performance at 86.4\% accuracy.

Our research advances these existing methods by introducing cutting-edge models like ResNet50, MobileNetV3, and EfficientNet-B2 specifically for ArSL recognition. Additionally, explainable AI (XAI) techniques have been incorporated to enhance model transparency, which is vital in sensitive fields like healthcare and education. Our preprocessing pipeline includes oversampling and extensive data augmentation, addressing diverse data scenarios more effectively. This system not only surpasses previous models in accuracy but also ensures adaptability and scalability, making it applicable to other sign languages.

A summary of these related works is presented in Table \ref{tab:relatedworks}, outlining the datasets, methodologies, and results achieved. This table offers a clear comparison of different approaches to Arabic sign language recognition, showcasing the diversity in techniques and performance levels.

\section{Datasets}\label{sec:dataset}

This study employs two datasets for Arabic alphabet sign language recognition: the ArSL2018 dataset \cite{latif2018arabic} and the RGB Arabic Alphabets Sign Language Dataset (AASL) \cite{https://doi.org/10.48550/arxiv.2301.11932}.

\subsection{Arabic Alphabets Sign Language Dataset (ArASL2018)}

\begin{figure}[h]
\centering
\includegraphics[width=0.15\textwidth]{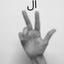}
\includegraphics[width=0.15\textwidth]{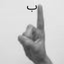}
\includegraphics[width=0.15\textwidth]{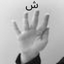}
\caption{Examples of images from the ArSL2018 dataset}
\label{fig:ArSL2018_images}
\end{figure}

The ArSL2018 dataset, introduced by Latif et al. \cite{latif2018arabic}, comprises 54,049 grayscale images, each sized at 64x64 pixels, representing 32 Arabic sign language signs and alphabets. The dataset was collected from 40 participants of diverse age groups in Al Khobar, Saudi Arabia, using an iPhone 6S camera. To enhance robustness, the dataset includes variations in lighting, angles, and backgrounds. Figure \ref{fig:ArSL2018_images} provides examples of these images.

\begin{figure}[h]
\centering
\includegraphics[width=0.4\textwidth]{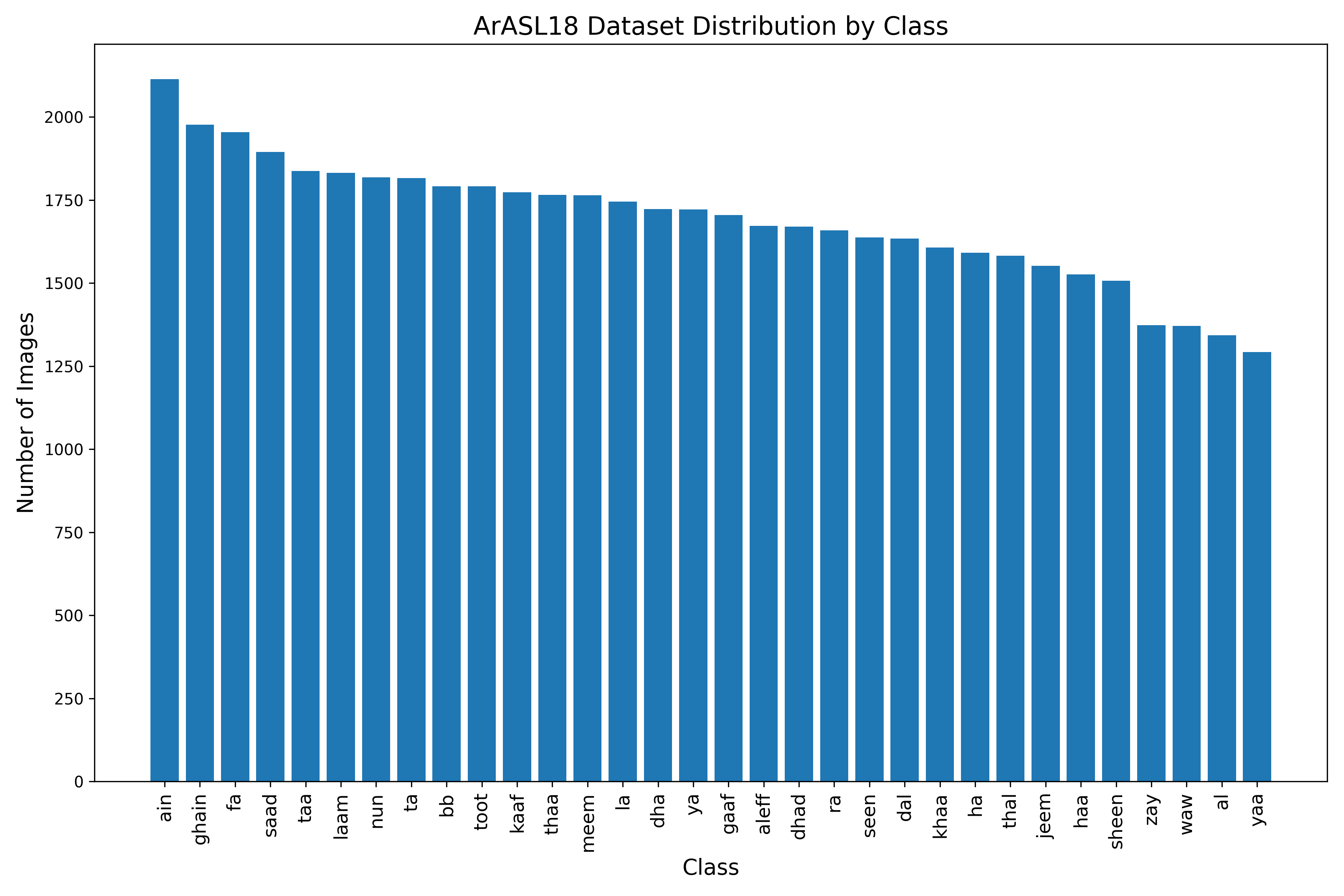}
\caption{Class distribution of the ArSL2018 dataset}
\label{fig:ArSL2018_distribution}
\end{figure}

The class distribution in ArSL2018, illustrated in Figure \ref{fig:ArSL2018_distribution}, shows an uneven number of samples across classes. This imbalance could introduce bias, potentially affecting the model's generalization performance.

\subsection{RGB Arabic Alphabets Sign Language Dataset (AASL)}

\begin{figure}[h]
\centering
\includegraphics[width=0.15\textwidth]{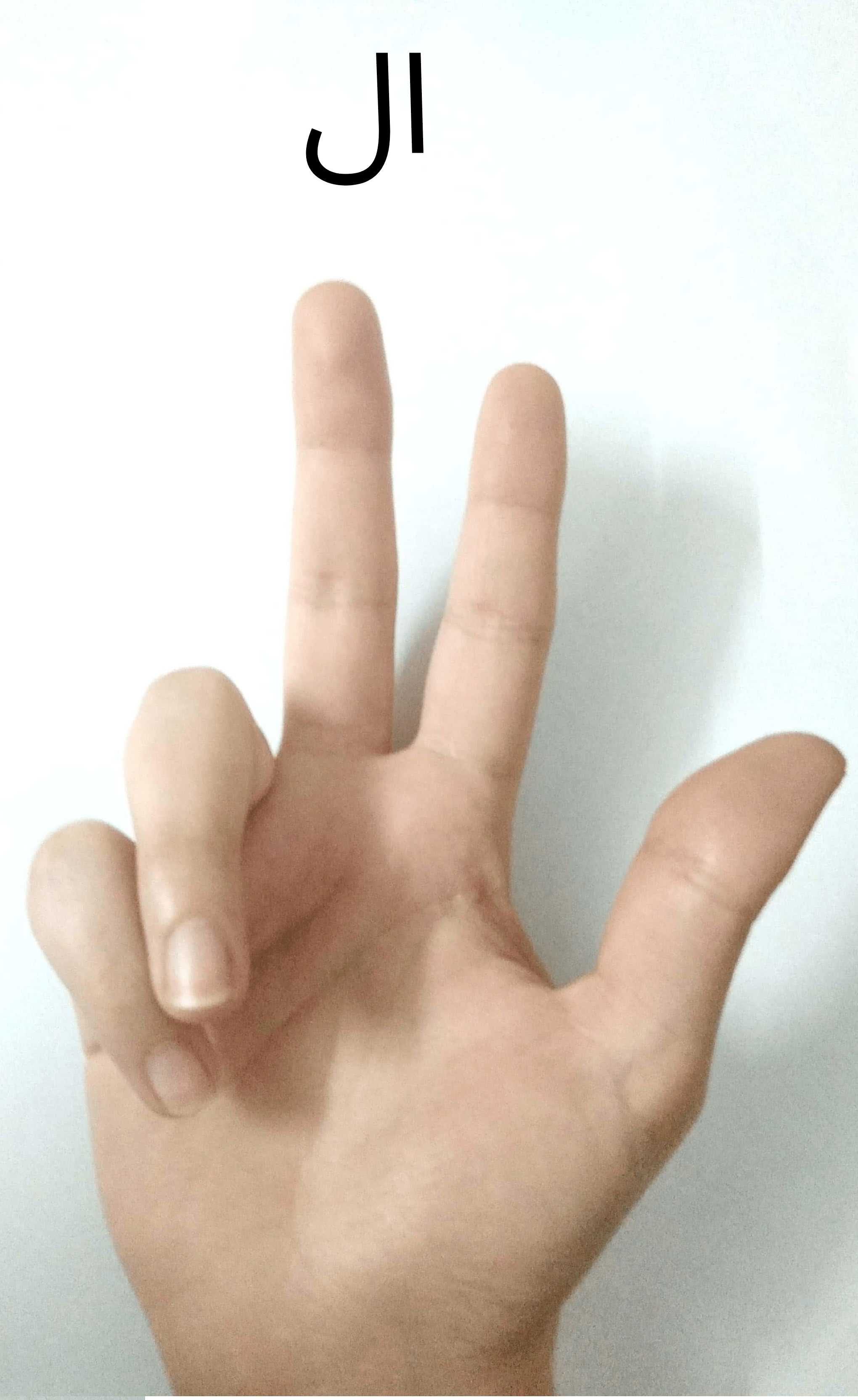}
\includegraphics[width=0.15\textwidth]{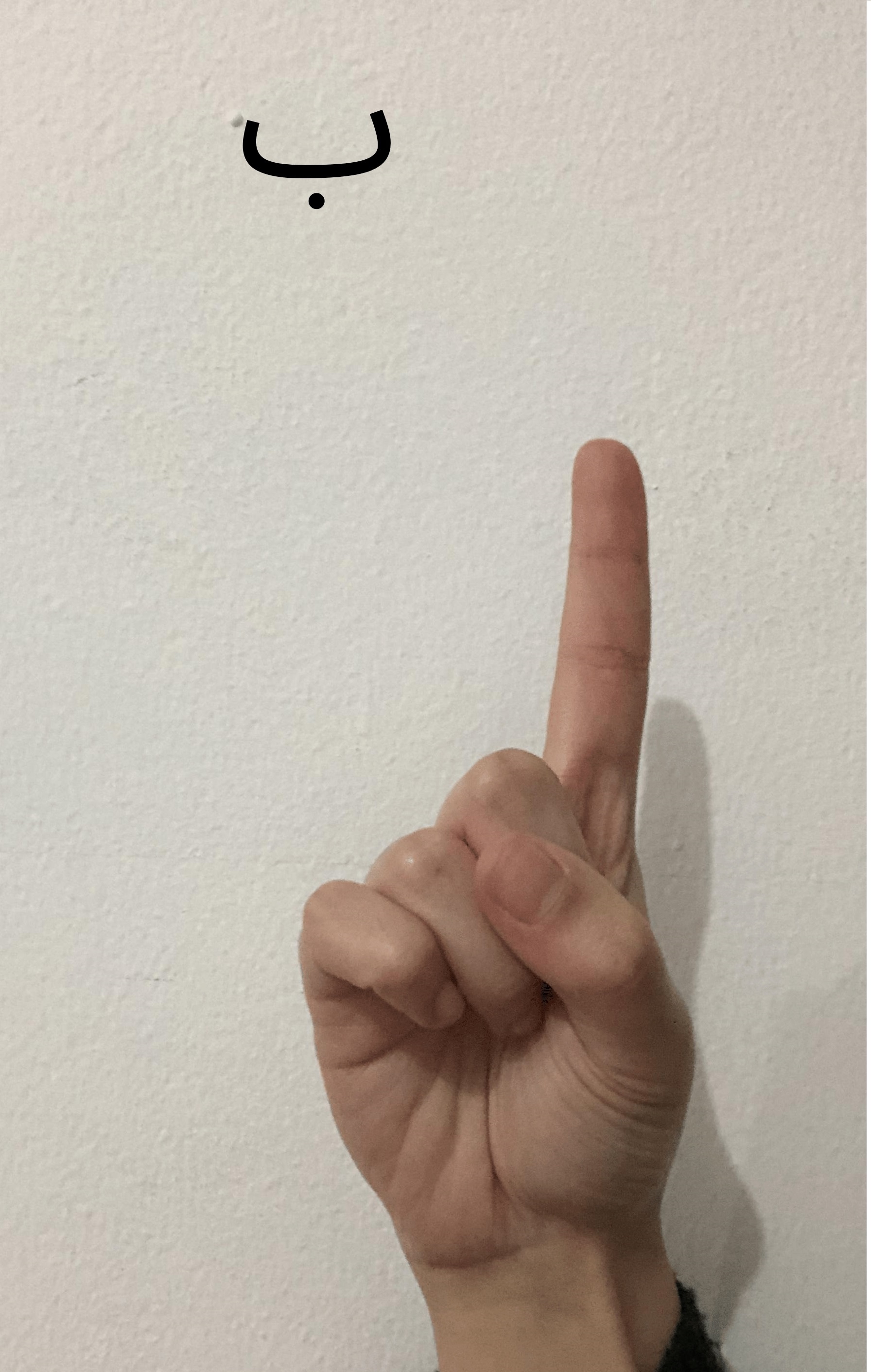}
\includegraphics[width=0.15\textwidth]{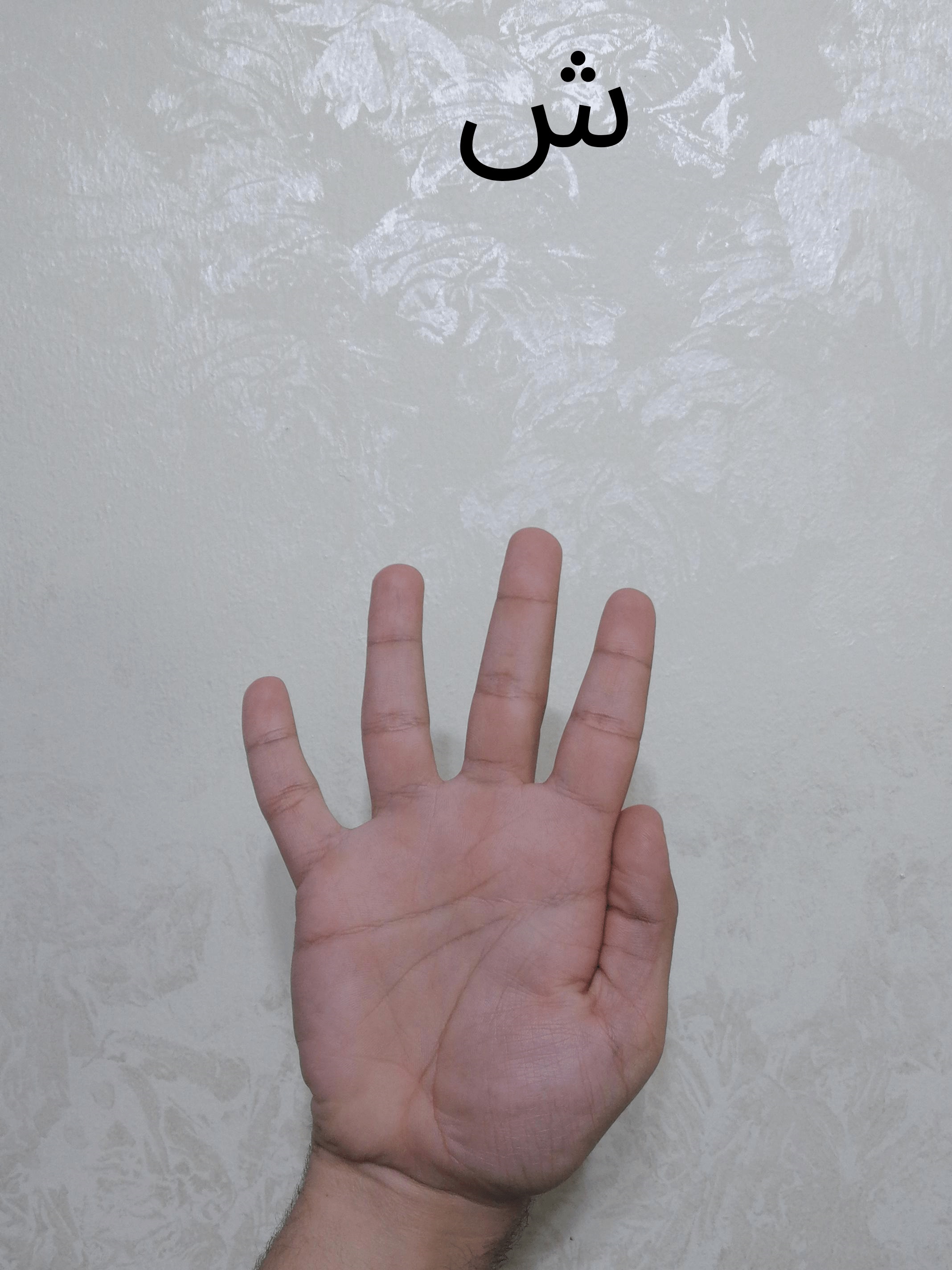}
\caption{Examples of images from the AASL dataset}
\label{fig:AASL_images}
\end{figure}

The AASL dataset \cite{https://doi.org/10.48550/arxiv.2301.11932} contains 7,857 labeled RGB images representing 31 Arabic sign language alphabets. Collected from over 200 participants using various types of cameras, the dataset includes a range of conditions, such as diverse lighting, backgrounds, and orientations, enhancing its suitability for real-world applications. Examples of these images are presented in Figure \ref{fig:AASL_images}.

\begin{figure}[h]
\centering
\includegraphics[width=0.4\textwidth]{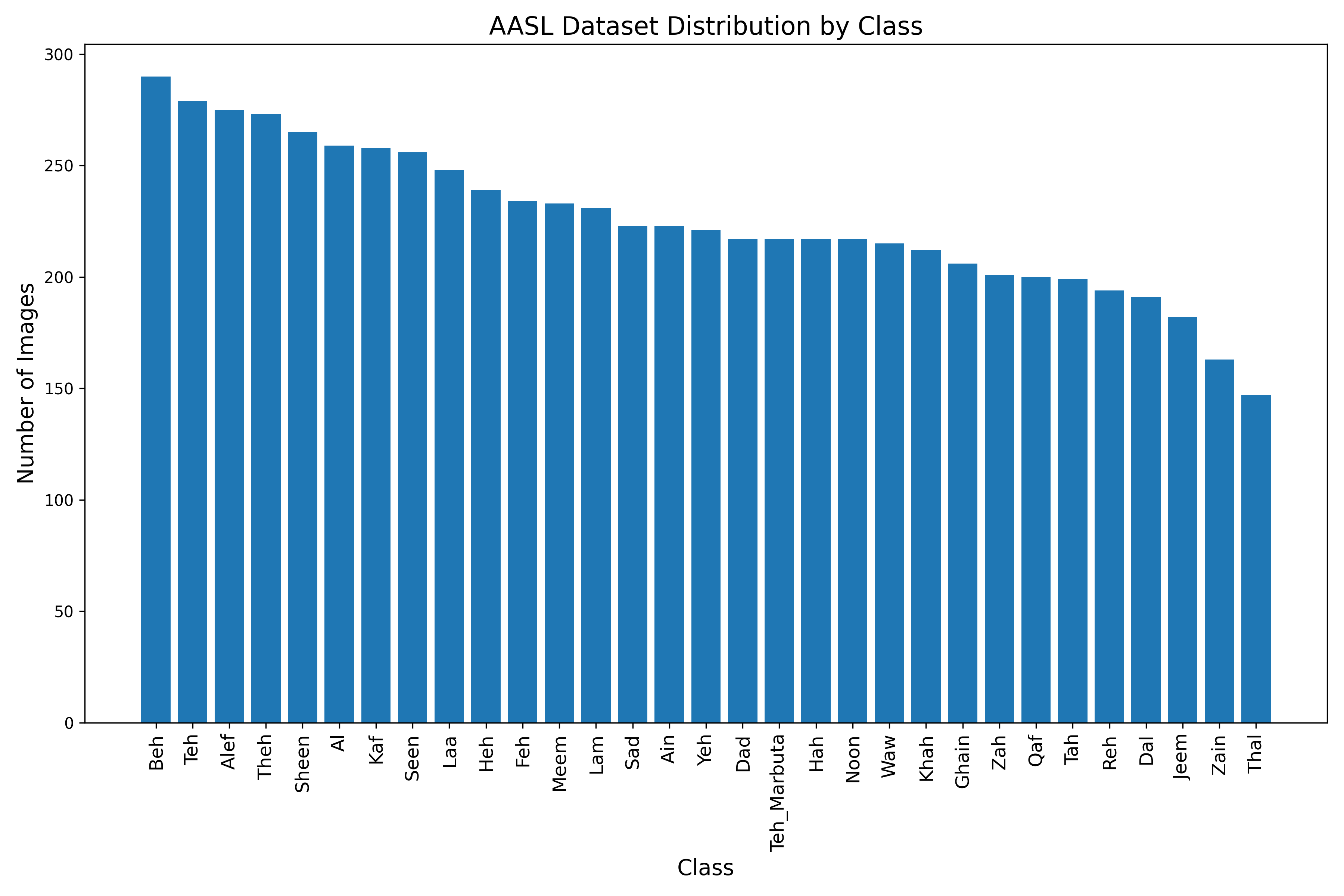}
\caption{Class distribution of the AASL dataset}
\label{fig:AASL_distribution}
\end{figure}

As seen in Figure \ref{fig:AASL_distribution}, the AASL dataset also exhibits uneven class distribution, which may skew model performance toward more represented classes.

\section{Methodology}
\label{sec:methodology}

\subsection{Data Preparation}

Ensuring a balanced and standardized dataset was essential for improving model performance and reducing bias. The data preparation phase focused on handling class imbalance and implementing preprocessing techniques to standardize the input data.

\subsection{Handling Class Imbalance: ArSL2018 and AASL Datasets }

\begin{figure}[h]
\centering
\includegraphics[width=0.4\textwidth]{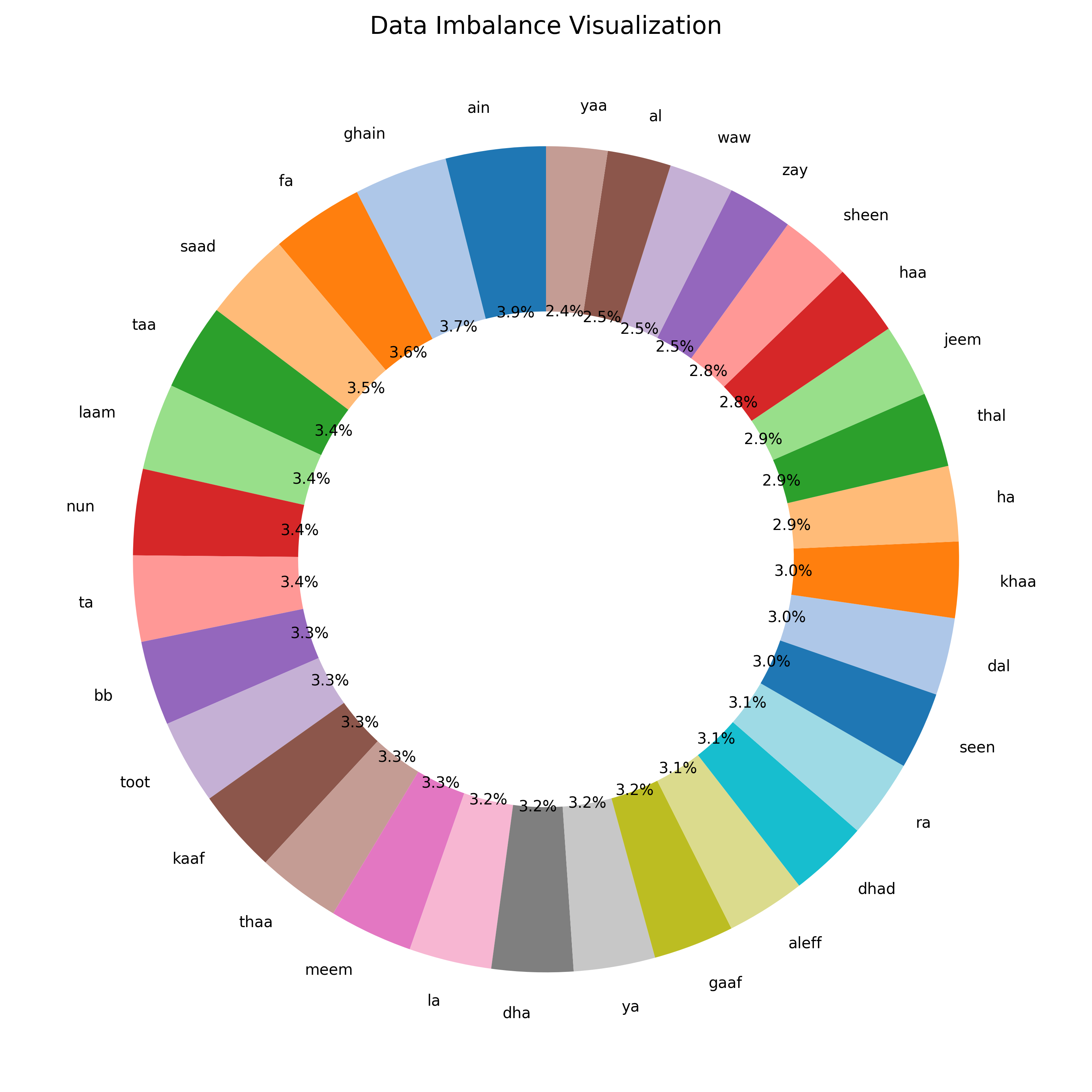}
\caption{Class imbalance ratio for ArSL2018 dataset}
\label{fig:ArSL2018_imbalance}
\end{figure}

Given the differences in class distribution, separate strategies were employed for the ArSL2018 and AASL datasets:

\textbf{ArSL2018 Dataset:}  The ArSL2018 dataset exhibited significant overrepresentation of certain signs, which could lead to model bias. To address this issue, undersampling was applied, capping the number of images at 1,250 per class. This adjustment balanced the dataset, reducing the likelihood of bias towards more frequent classes. Figure \ref{fig:ArSL2018_imbalance} displays the distribution after undersampling, confirming balanced representation across classes. This step was crucial for enhancing model generalization and ensuring equitable learning.

\textbf{AASL Dataset:} In contrast, due to the smaller sample size in the AASL dataset, no undersampling was performed. All samples were preserved to maintain sufficient training data, ensuring that the model could capture the dataset’s full diversity without compromising performance.

\begin{figure}[h]
\centering
\includegraphics[width=0.4\textwidth]{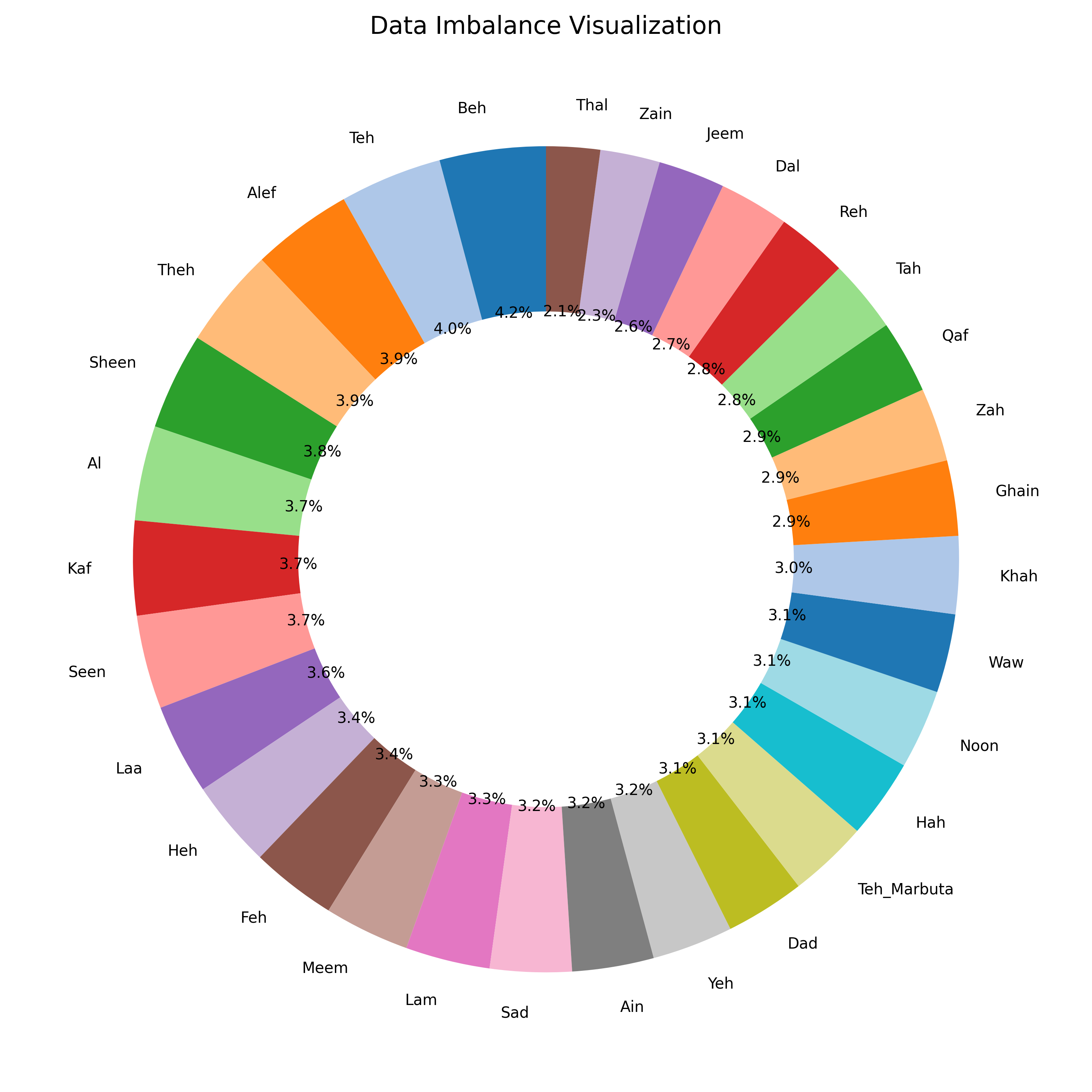}
\caption{Class imbalance ratio for AASL dataset}
\label{fig:AASL_imbalance}
\end{figure}

\subsection{Preprocessing}

The preprocessing steps were designed to standardize the datasets while ensuring consistency for model training.

\begin{itemize}
    \item \textbf{Image Resizing:}
    Images from both datasets were resized to \textbf{224x224 pixels} to match the input requirements of deep learning models like MobileNet, ResNet, and EfficientNet. This resizing ensured compatibility across models and uniform feature extraction. Bilinear interpolation was used for resizing, offering a balance in image quality by averaging pixel values \cite{rukundo2023effects}.

    \item \textbf{Normalization:}
    After resizing, pixel values were scaled to the range \textbf{[0, 1]} by dividing by 255. Additionally, standardization was applied by adjusting pixel values to have a mean of zero and a standard deviation of one, based on the training dataset. This step aligned the inputs with models pre-trained on ImageNet, which assume standardized inputs \cite{lima2023large}.
\end{itemize}

These preprocessing steps were crucial for improving model convergence, maintaining uniform scaling, and ensuring consistent data preparation across both datasets.

\begin{figure}[h]
\centering
\includegraphics[width=7cm,height=5cm]{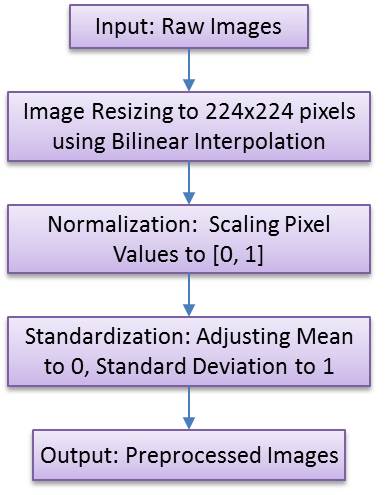}
\caption{Flowchart of the preprocessing pipeline.}
\label{fig:preprocessing_flowchart}
\end{figure}

%%%%%%%%%%%%%%%%%%%%%%%%%%%%%%%%%%%%

As depicted in Figure~\ref{fig:preprocessing_flowchart}, the preprocessing pipeline converts raw images into a standardized format suitable for deep learning models.

\subsection{Data Splitting}

To facilitate effective model training, tuning, and evaluation, the datasets were split into training, validation, and test sets using an \textbf{80:10:10 ratio}. This approach ensured balanced data allocation across subsets, supporting generalization while preserving unbiased evaluation \cite{Joseph_2022}.

\textbf{Training Set (80\%)}
The training set contained 80\% of the total data, allowing models to learn patterns and relationships through multiple forward and backward passes. A larger data allocation enabled the models to generalize effectively across diverse features.

\textbf{Validation Set (10\%)}
The validation set comprised 10\% of the data, used during training to tune hyperparameters and evaluate model performance after each epoch. It provided an intermediate assessment, helping prevent overfitting and facilitating adjustments in parameters like learning rate, batch size, and architecture.

\textbf{Test Set (10\%)}
The test set represented the remaining 10\% and was reserved exclusively for final model evaluation. As it remained unseen during training, it provided an unbiased estimate of the model’s real-world performance, ensuring reliable results for accuracy, precision, recall, and other metrics.

\subsection{Stratified Splitting}

To maintain consistent class distribution across training, validation, and test sets, \textbf{stratified splitting} was applied. This technique preserved class proportions in each subset, reducing bias towards specific classes during evaluation \cite{khan2022balancedsplitnewtraintest}. Stratified splitting thus enhanced the robustness of model evaluation and ensured consistent metrics across all subsets.

The data splitting process was critical for developing a reliable and generalizable model, enabling it to handle unseen data effectively while maintaining consistent performance across evaluation stages.

%%%%%%%%%%%%%%%%%%%%%%%%%%%%%%%%%%%%%%%%%%

\section{Model Training and Description}
\label{sec:model}
Model training was a multi-step process that involved selecting suitable models, defining a well-structured training procedure, and optimizing hyperparameters to ensure robust performance across evaluation metrics. Three models were chosen for this study: MobileNetV3 \cite{koonce2021mobilenetv3}, ResNet50 \cite{mascarenhas2021comparison}, and EfficientNet-B2 \cite{zhao2024corrosion}, each selected based on its ability to handle the complexity of the dataset while balancing performance and computational efficiency.

\subsection{Model Selection}
\begin{figure}[h]
    \centering
    \includegraphics[width=0.4\textwidth]{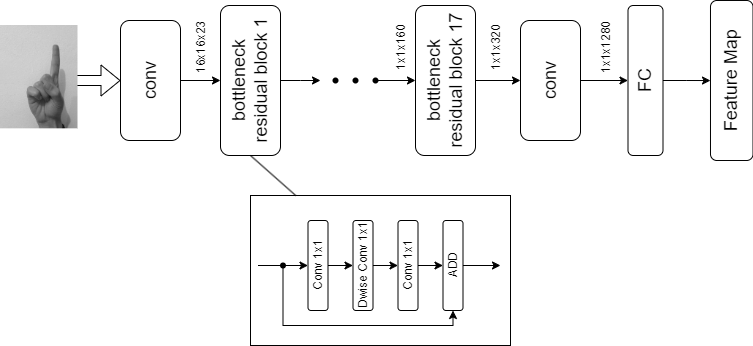}
    \caption{MobileNetV3 Architecture, featuring depthwise separable convolutions and SE blocks for improved performance and efficiency.}
    \label{fig:mobilenet_architecture}
\end{figure}

Each model used in this study has distinct advantages that make it well-suited for Arabic sign language recognition:

\paragraph{MobileNetV3}
MobileNetV3 \cite{koonce2021mobilenetv3} is a lightweight convolutional neural network designed for efficiency and speed, making it suitable for real-time applications. It uses depthwise separable convolutions to reduce computational cost while maintaining high accuracy. The key equation in MobileNetV3’s architecture is:

\begin{equation}
y_i = \text{ReLU6} \left( \sum_{j=1}^{k} \text{DWConv}(w_{ij} * x_j) + b_i \right),
\end{equation}

where \( \text{DWConv} \) represents depthwise convolution, \( w_{ij} \) are the weights, \( x_j \) are the input feature maps, and \( b_i \) is the bias term. Here, \( k \) is the number of hidden layers. The ReLU6 activation function \cite{lee2023gelu} is a variant of the ReLU function that caps the output at 6, which helps in preventing the activation from becoming too large. Figure~\ref{fig:mobilenet_architecture} shows the architecture of MobileNetV3, highlighting its use of squeeze-and-excitation (SE) blocks and inverted residuals \cite{nguyen2024hyperspectral}.

\paragraph{ResNet50}
ResNet50 \cite{mascarenhas2021comparison} is known for its deep residual learning approach \cite{hu2024spiking}, which solves the vanishing gradient problem by introducing residual connections. The core equation for a residual block in ResNet50 is:

\begin{equation}
y = x + F(x, \{W_i\}),
\end{equation}

where \( x \) is the input, \( F(x, \{W_i\}) \) represents the residual mapping, and \( y \) is the output. Here, \( \{W_i\} \) denotes the set of weights for the layers within the residual block. This architecture allows the network to learn identity mappings more easily, improving convergence and enabling the training of much deeper networks.

The residual block in ResNet50 typically consists of two or three convolutional layers with batch normalization and ReLU activation functions \cite{zhang2024deep}. The key innovation is the addition of a shortcut connection that skips one or more layers, directly connecting the input to the output. This shortcut connection ensures that the gradient can flow directly through the network, mitigating the vanishing gradient problem.

The architecture of ResNet50 is designed to capture complex patterns and features in images, making it highly effective for image classification tasks. The use of residual connections not only improves the training of deep networks but also enhances the network's ability to generalize to new data.

\begin{figure}[h]
    \centering
    \includegraphics[width=0.4\textwidth,  height=.2\textwidth]{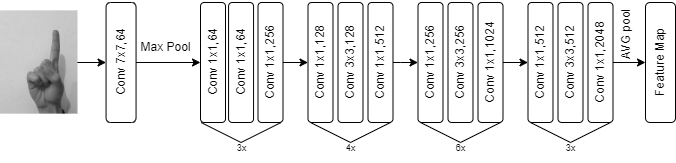}
    \caption{ResNet50 Architecture, highlighting the use of residual blocks to enable deeper networks with better convergence.}
    \label{fig:resnet_architecture}
\end{figure}

Figure~\ref{fig:resnet_architecture} presents the ResNet50 architecture, emphasizing its residual connections and the flow of information through the network.

\paragraph{EfficientNet-B2}
EfficientNet-B2 \cite{zhao2024corrosion} is part of the EfficientNet family of models, designed to achieve high accuracy with fewer parameters and lower computational cost. The EfficientNet models use a compound scaling method that uniformly scales the depth, width, and resolution of the network. The scaling is governed by the following equations:

\begin{equation}
d = \alpha^\phi,~~~~ w = \beta^\phi, ~~~~   r = \gamma^\phi,
\end{equation}

where \( d \) is the depth, \( w \) is the width, \( r \) is the resolution, and \( \alpha, \beta, \gamma \) are constants determined through a grid search. The parameter \( \phi \) is a compound coefficient that controls the scaling of the network.

The EfficientNet architecture is based on the Mobile Inverted Residual Bottleneck Block (MBConv) \cite{shang2023high}, which incorporates depthwise separable convolutions, squeeze-and-excitation (SE) blocks, and swish activation functions \cite{rahman2024swishrelu}. These components contribute to the model's efficiency and performance.

EfficientNet-B2 specifically balances performance and computational efficiency, making it suitable for a wide range of applications. It achieves this by carefully scaling the network dimensions while maintaining a balance between model complexity and computational resources.

\begin{figure}[h]
    \centering
    \includegraphics[width=0.4\textwidth ,  height=.2\textwidth]{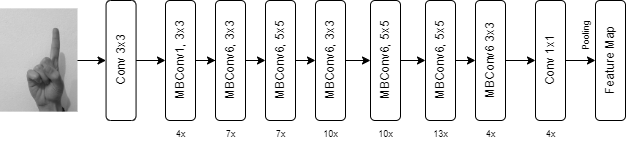}
    \caption{EfficientNet-B2 Architecture, using compound scaling to balance network depth, width, and resolution for optimal performance.}
    \label{fig:efficientnet_architecture}
\end{figure}

Figure~\ref{fig:efficientnet_architecture} illustrates the EfficientNet-B2 architecture, demonstrating its compound scaling approach. The architecture shows how the depth, width, and resolution are scaled uniformly, allowing the model to capture complex patterns in the data while remaining computationally efficient.

\subsection{Training Procedure}
To ensure robustness and generalizability, \textbf{5-Fold Cross-Validation} was employed. This method divides the data into five subsets (folds), with four used for training and one for validation in each iteration. The process is repeated five times, with each fold serving as the validation set once. The final performance metrics are obtained by averaging the results across all folds, minimizing bias and variance.

This cross-validation approach provided a comprehensive evaluation of model performance across different data partitions, reducing the risk of overfitting and increasing confidence in the model’s generalization to unseen data.

\subsection{Hyperparameters}
The models were trained using a set of optimized hyperparameters to ensure effective learning and prevent overfitting. Table~\ref{tab:hyperparameters} outlines the hyperparameters used for training.

\begin{table}[h]
    \centering
    \caption{Hyperparameters used for training the models}
    \label{tab:hyperparameters}
    \begin{tabular}{|>{\centering\arraybackslash}m{3cm}|>{\centering\arraybackslash}m{3cm}|}
        \hline
        \textbf{Hyperparameter} & \textbf{Value} \\
        \hline
        Optimizer & AdamW \\
        Learning Rate & 0.0001 \\
        Batch Size & 16 \\
        Number of Epochs & 10 \\
        Early Stopping & Enabled \\
        Learning Rate Scheduler & ReduceLROnPlateau \\
        Weight Decay & 0.0001 \\
        \hline
    \end{tabular}
\end{table}

% Early stopping was implemented to prevent overfitting by halting training when the validation loss ceased to improve over a set number of epochs (patience). Additionally, a learning rate scheduler (\texttt{ReduceLROnPlateau}) \cite{thakur2024transformative} was used to reduce the learning rate when a plateau in validation performance was detected, facilitating better convergence.

Early stopping was implemented to prevent overfitting by halting training when the validation loss ceased to improve over a set number of epochs (patience). This technique helps in avoiding unnecessary training and ensures that the model does not become too specialized to the training data, thereby improving its generalization to unseen data.

Additionally, a learning rate scheduler (ReduceLROnPlateau) \cite{thakur2024transformative} was used to reduce the learning rate when a plateau in validation performance was detected. This scheduler dynamically adjusts the learning rate during training, reducing it by a specified factor when the validation loss has not improved for a certain number of epochs.

\begin{table}[h]
\centering
\caption{Model Training and Evaluation Workflow}
\label{tab:workflow}
\textbf{Input:} Raw images from ArSL2018 and AASL datasets. \\
\textbf{Output:} Trained models (MobileNetV3, ResNet50, EfficientNet-B2) with performance metrics. \\

The following steps describe the model training and evaluation process:

\begin{enumerate}
    \item \textbf{Data Preparation}
    \begin{itemize}
        \item Handle class imbalance using undersampling for ArSL2018 and preserving all samples for AASL.
        \item Implement preprocessing techniques including image resizing and normalization.
    \end{itemize}

    \item \textbf{Preprocessing}
    \begin{itemize}
        \item Resize images to 224x224 pixels using bilinear interpolation.
        \item Normalize pixel values to the range [0, 1].
        \item Standardize pixel values using mean and standard deviation from the training dataset.
    \end{itemize}

    \item \textbf{Data Splitting}
    \begin{itemize}
        \item Divide datasets into training (80\%), validation (10\%), and test (10\%) sets.
        \item Apply stratified splitting to maintain consistent class distribution across subsets.
    \end{itemize}

    \item \textbf{Model Selection}
    \begin{itemize}
        \item Choose MobileNetV3 \cite{abd2023evolution}, ResNet50 \cite{mukherjee2023annotatedresnet}, and EfficientNet-B2 \cite{agarwal2020efficientnet}based on their suitability for the task.
    \end{itemize}

    \item \textbf{Training Procedure}
    \begin{itemize}
        \item Employ 5-Fold Cross-Validation to ensure robustness and generalizability.
        \item Train models using optimized hyperparameters.
    \end{itemize}

    \item \textbf{Hyperparameter Optimization}
    \begin{itemize}
        \item Use AdamW optimizer with a learning rate of 0.0001.
        \item Implement early stopping and learning rate scheduler (ReduceLROnPlateau).
    \end{itemize}

    \item \textbf{Model Evaluation}
    \begin{itemize}
        \item Evaluate model performance using accuracy, precision, recall, and other metrics.
        \item Document the results and performance metrics.
    \end{itemize}

    \item \textbf{Model Saving}
    \begin{itemize}
        \item Save the trained models and their parameters for future use.
    \end{itemize}
\end{enumerate}
\end{table}

\begin{figure}[h!]
    \centering
    \includegraphics[width=9cm,height=6cm]{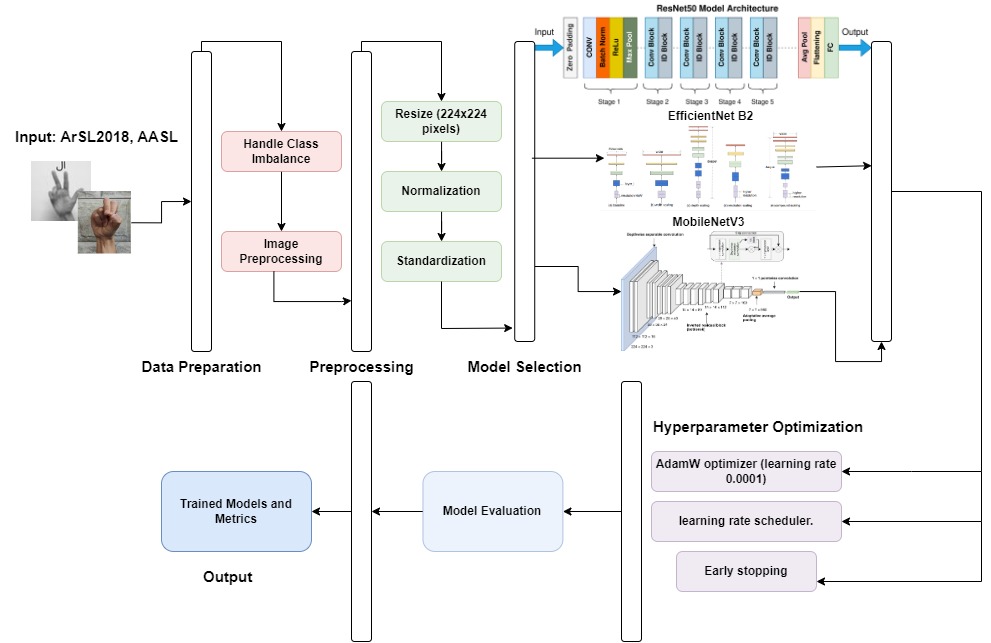}
    \caption{The proposed model framework}
    \label{fig:workflow}
\end{figure}
As shown in Fig. \ref{fig:workflow}, the proposed model framework outlines the steps for training and evaluating models. This systematic approach includes data preparation and preprocessing, data splitting, model selection, and training. The use of 5-Fold Cross-Validation and optimized hyperparameters ensures robust and generalizable models.

\section{Performance Evaluation Metrics}
\label{sec:evaluation}
To provide a comprehensive assessment of the models’ performance, multiple evaluation metrics were used, including accuracy, F1-score, precision, and recall. These metrics offer insights into different aspects of model performance, particularly in the context of class imbalance.

\textbf{Accuracy}
Accuracy measures the proportion of correctly classified instances among the total number of instances. It is a straightforward metric, but it may not always be reliable in the presence of class imbalance, as it could be biased towards the majority class. The formula for accuracy is:

\begin{equation}
\text{Accuracy} = \frac{TP + TN}{TP + TN + FP + FN},
\end{equation}

where:
\begin{itemize}
    \item \( TP \) = True Positives (correctly predicted positive samples)
    \item \( TN \) = True Negatives (correctly predicted negative samples)
    \item \( FP \) = False Positives (incorrectly predicted positive samples)
    \item \( FN \) = False Negatives (incorrectly predicted negative samples)
\end{itemize}

While accuracy provides a general overview of how well the model performs, it may not fully capture performance in imbalanced datasets, as it does not differentiate between the types of errors made by the model.

\textbf{Precision}
Precision focuses on the correctness of positive predictions, making it an important metric when false positives are costly. It indicates the proportion of true positive predictions among all positive predictions made by the model. The formula for precision is:

\begin{equation}
\text{Precision} = \frac{TP}{TP + FP}.
\end{equation}

Precision is particularly useful in applications where minimizing false positives is crucial, such as in medical diagnoses or accident detection, where predicting a positive class incorrectly could have serious consequences.

\textbf{Recall}
Recall, also known as sensitivity or true positive rate, measures the model’s ability to identify all relevant instances. It represents the proportion of true positive samples that were correctly identified out of the total actual positive samples. The formula for recall is:

\begin{equation}
\text{Recall} = \frac{TP}{TP + FN}.
\end{equation}

Recall is critical in scenarios where minimizing false negatives is more important, such as detecting accidents or medical conditions where missing a positive case could have severe implications.

\textbf{F1-Score}
The F1-score is the harmonic mean of precision and recall, providing a single measure that balances both metrics. It is particularly useful when dealing with imbalanced datasets, as it accounts for both false positives and false negatives. The formula for the F1-score is:

\begin{equation}
\text{F1-Score} = 2 \cdot \frac{\text{Precision} \cdot \text{Recall}}{\text{Precision} + \text{Recall}}.
\end{equation}

An F1-score close to 1 indicates a good balance between precision and recall, making it effective in scenarios where both false positives and false negatives need to be minimized. It is a robust metric for evaluating models in the context of class imbalance, as it considers both over-prediction and under-prediction of classes.

\textbf{Interpretation and Use}
The choice of evaluation metrics is based on the specific requirements of the task and the dataset characteristics:

\begin{itemize}
    \item \textbf{Accuracy:} Provides a general overview of model performance but may not be sufficient in cases of severe class imbalance.
    \item \textbf{Precision:} Important when false positives need to be minimized.
    \item \textbf{Recall:} Critical in cases where false negatives are more detrimental.
    \item \textbf{F1-Score:} Offers a balanced evaluation, especially useful for imbalanced datasets.
\end{itemize}

The combination of these metrics ensures a comprehensive evaluation of the model’s performance, addressing both the correctness of predictions and the impact of class imbalance. This allows for better decision-making in model selection and fine-tuning.

\section{Explainable AI}
\label{sec:XAI}

To enhance the transparency and interpretability of the model’s predictions, Explainable AI (XAI) techniques were integrated. Among these, \textbf{Grad-CAM (Gradient-weighted Class Activation Mapping)} was used to provide visual insights into the model's decision-making process. This approach helps users understand which areas of the input data influenced the model's predictions, improving trust and enabling more informed decision-making.

\subsection{Importance of Explainable AI}
Explainable AI is critical in real-world applications where understanding model predictions is as important as achieving high accuracy. The benefits of XAI include:

\begin{itemize}
    \item \textbf{Transparency:} It allows stakeholders to identify which parts of the input data had the greatest influence on model decisions.
    \item \textbf{Trust and Reliability:} It builds user confidence by making the AI's decision-making process clearer, which is particularly crucial in sensitive fields like healthcare, accident detection, and autonomous systems.
    \item \textbf{Error Analysis:} By analyzing why a model made incorrect predictions, XAI helps identify biases or errors, aiding in debugging and model refinement.
    \item \textbf{Compliance:} It meets ethical and regulatory requirements, ensuring accountability in fields that demand responsible AI use, such as finance, healthcare, and law.
\end{itemize}

\subsection{Grad-CAM Overview}
Grad-CAM is a powerful tool for visualizing which regions of an input image contribute most to the model’s predictions. It generates heatmaps that highlight the areas relevant to a specific decision, making the model's behavior easier to interpret. The core steps of Grad-CAM are outlined below.

\paragraph{Grad-CAM Methodology}
Grad-CAM works by calculating the gradient of the predicted class score concerning feature maps from a convolutional layer. These gradients are then used to create a weighted combination of the feature maps, focusing on the most critical regions. The main steps are:

\begin{enumerate}
    \item \textbf{Calculate the gradients:} Compute the gradient of the target class score \( y_c \) with respect to the feature maps \( A^k \) from a selected convolutional layer:
    \begin{equation}
    \frac{\partial y_c}{\partial A^k}.
    \end{equation}

    \item \textbf{Compute weights:} The gradients are global average-pooled over the width and height of the feature maps to obtain weights \( \alpha_k \) for each feature map:
    \begin{equation}
    \alpha_k = \frac{1}{Z} \sum_i \sum_j \frac{\partial y_c}{\partial A_{ij}^k},
    \end{equation}
    where \( Z \) is the total number of pixels in the feature map.

    \item \textbf{Generate the Grad-CAM heatmap:} The final heatmap \( L_{\text{Grad-CAM}} \) is obtained by taking a weighted sum of the feature maps, followed by a ReLU activation to focus on the most relevant regions:
    \begin{equation}
    L_{\text{Grad-CAM}} = \text{ReLU} \left( \sum_k \alpha_k A^k \right).
    \end{equation}
\end{enumerate}

\begin{figure}[t]
    \centering
    \includegraphics[width=0.4\textwidth]{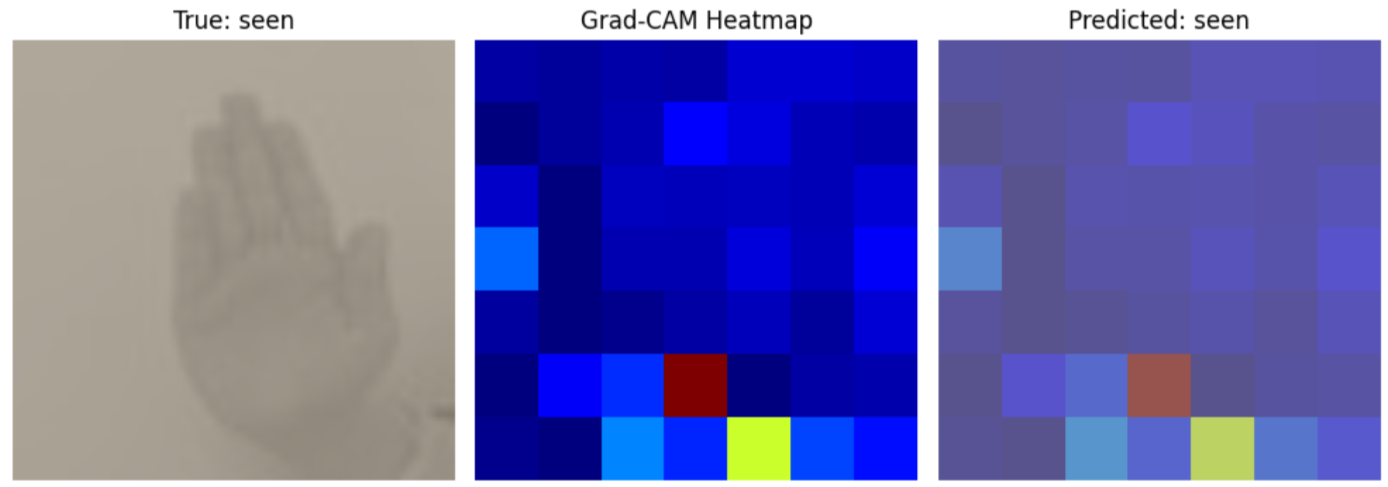}
    \caption{Example of a Grad-CAM heatmap superimposed on an input image, showing regions most relevant to the model’s prediction.}
    \label{fig:grad_cam_example}
\end{figure}

\paragraph{Visual Insights from Grad-CAM}
Grad-CAM produces a heatmap that can be overlaid on the original input image to highlight areas most influential in the model's decision-making. This visualization gives users an intuitive understanding of the image regions crucial for a particular prediction. Figure~\ref{fig:grad_cam_example} illustrates an example of a Grad-CAM heatmap.

\subsection{Application of Grad-CAM in Model Evaluation}
In this study, Grad-CAM was applied to models like MobileNet, ResNet, and EfficientNet to analyze their predictions. The resulting heatmaps provided the following insights:

\begin{itemize}
    \item \textbf{Identifying Biases:} Grad-CAM visualizations revealed potential biases, such as over-reliance on specific image features or irrelevant regions, guiding model refinement.
    \item \textbf{Assessing Model Reliability:} Consistent focus on relevant image regions across samples indicated reliable performance, whereas inconsistent patterns suggested areas for further improvement.
    \item \textbf{Improving Trust in AI Decisions:} The visual explanations provided by Grad-CAM helped users, especially non-experts, understand model behavior, fostering trust and acceptance of AI decisions.
\end{itemize}

%%%%%%%%%%%%%%%%%%%%%%%%%%%%%%%%%%%%%%%%
\section{MobileNetV3 Results}\label{sec:resultsMobileNet3}

This section presents the experimental results of the MobileNetV3 model on the ArSL2018 and AASL datasets. The results are divided into subsections for each dataset. For each model, we provide validation and test results, followed by analysis using confusion matrices and Grad-CAM to interpret the models' decision-making.

\subsection{ArSL2018 Dataset}
In this subsection, we present the results of our experiments on the ArSL2018 dataset using MobileNetV3 model. The results are reported for both validation and test sets, along with confusion matrices and Grad-CAM visualizations to interpret model performance.
%\subsubsection{MobileNetV3}
\textbf{Validation Results:} The validation results for MobileNetV3 on the ArSL2018 dataset are summarized in Table~\ref{tab:mobilev3_arsl2018_val}, showing high F1-Scores across all folds, with Fold 5 being the best.

\begin{table}[h]
    \centering
    \caption{Validation Results for MobileNetV3 (ArSL2018 Dataset)}
    \label{tab:mobilev3_arsl2018_val}
    \begin{tabular}{|c|c|c|c|c|}
        \hline
        \textbf{Fold} & \textbf{Precision} & \textbf{Recall} & \textbf{F1-Score} & \textbf{Accuracy} \\
        \hline
        Fold 1 & 0.994899 & 0.994862 & 0.994859 & 0.994844 \\
        Fold 2 & 0.996274 & 0.996197 & 0.996218 & 0.996250 \\
        Fold 3 & 0.998295 & 0.998241 & 0.998257 & 0.998281 \\
        Fold 4 & 0.998162 & 0.998078 & 0.998106 & 0.998125 \\
        Fold 5 & 0.998592 & 0.998556 & 0.998556 & 0.998594 \\
        \hline
    \end{tabular}
\end{table}

\textbf{Test Results:} The test results for MobileNetV3, shown in Table~\ref{tab:mobilev3_arsl2018_test}, indicate consistent performance across all folds, with the highest F1-Score in Fold 5.

\begin{table}[h]
    \centering
    \caption{Test Results for MobileNetV3 (ArSL2018 Dataset)}
    \label{tab:mobilev3_arsl2018_test}
    \begin{tabular}{|c|c|c|c|c|}
        \hline
        \textbf{Fold} & \textbf{Precision} & \textbf{Recall} & \textbf{F1-Score} & \textbf{Accuracy} \\
        \hline
        Fold 1 & 0.986190 & 0.985611 & 0.985780 & 0.985750 \\
        Fold 2 & 0.986811 & 0.986502 & 0.986517 & 0.986500 \\
        Fold 3 & 0.989406 & 0.989344 & 0.989301 & 0.989250 \\
        Fold 4 & 0.991019 & 0.990715 & 0.990795 & 0.990750 \\
        Fold 5 & 0.992413 & 0.992222 & 0.992237 & 0.992250 \\
        \hline
    \end{tabular}
\end{table}

\begin{figure}[h]
    \centering
    \includegraphics[width=0.4\textwidth]{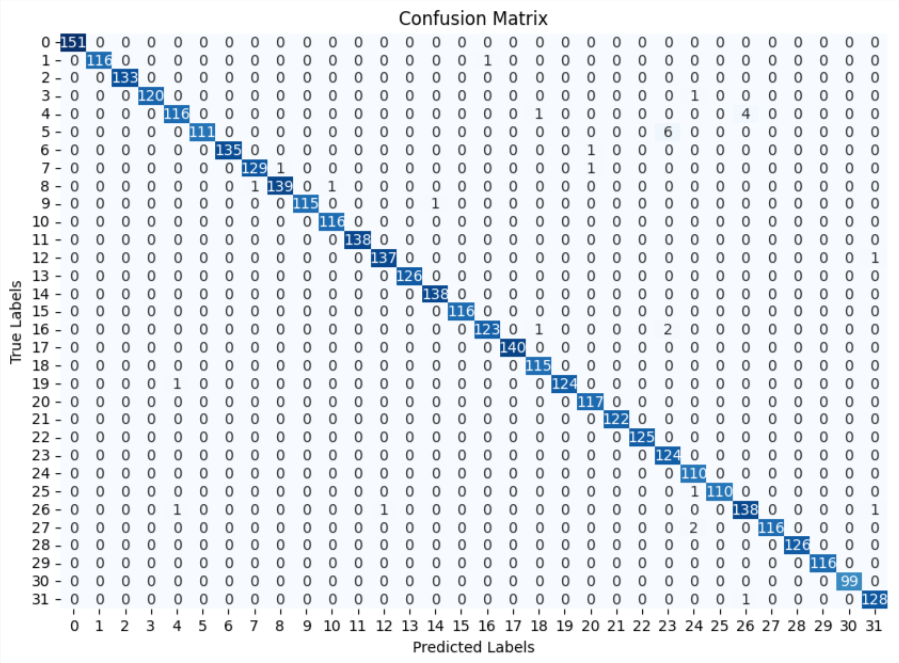}
    \caption{Confusion Matrix for MobileNetV3 (ArSL2018 Test Set, Fold 5)}
    \label{fig:cm_mobilev3_arsl2018}
\end{figure}

\textbf{Confusion Matrix:} The confusion matrix for MobileNetV3 (Fold 5) on the ArSL2018 test set is shown in Figure~\ref{fig:cm_mobilev3_arsl2018}. It indicates strong classification performance, with minimal misclassifications.

\begin{figure}[h]
    \centering
    \includegraphics[width=0.4\textwidth]{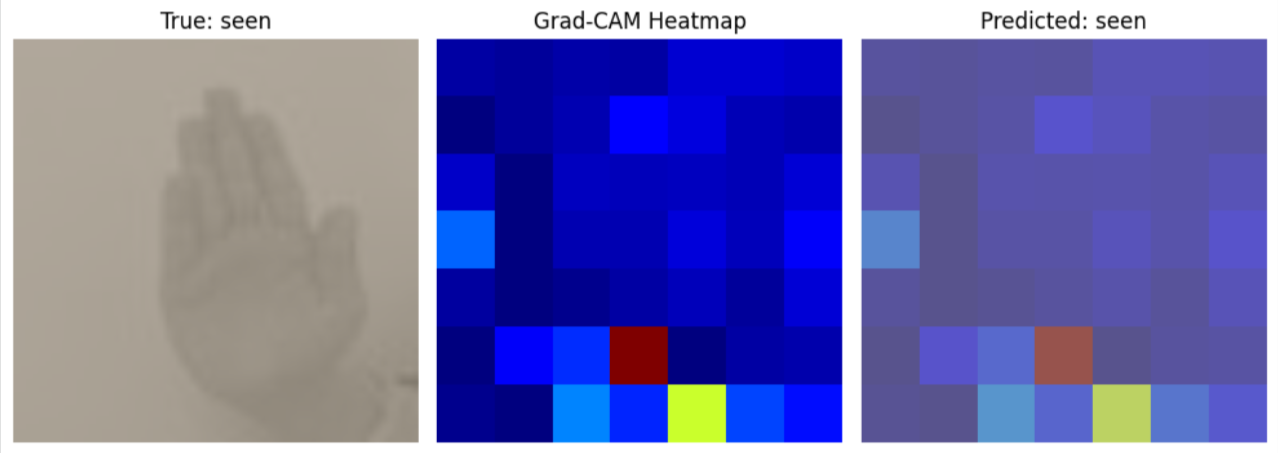}
    \caption{Grad-CAM for MobileNetV3 (ArSL2018 Test Set, Fold 5)}
    \label{fig:gradcam_mobilev3_arsl2018}
\end{figure}

\textbf{Grad-CAM Analysis:} Grad-CAM visualization for MobileNetV3 highlights the regions of the input images that contributed most to the model’s predictions. As shown in Figure~\ref{fig:gradcam_mobilev3_arsl2018}, the model focuses primarily on hand shapes, confirming its effectiveness in identifying relevant features for ArSL recognition.

\subsection{AASL Dataset}
%\subsubsection{MobileNetV3}
In this subsection, we present the results of our experiments on the AASL dataset using MobileNetV3 model. The results are reported for both validation and test sets, along with confusion matrices and Grad-CAM visualizations to interpret model performance.
\textbf{Validation Results:} The validation results for MobileNetV3 on the AASL dataset are shown in Table~\ref{tab:mobilev3_aasl_val}, with Folds 2, 4, and 5 achieving perfect scores.

\begin{table}[h]
    \centering
    \caption{Validation Results for MobileNetV3 (AASL Dataset)}
    \label{tab:mobilev3_aasl_val}
    \begin{tabular}{|c|c|c|c|c|}
        \hline
        \textbf{Fold} & \textbf{Precision} & \textbf{Recall} & \textbf{F1-Score} & \textbf{Accuracy} \\
        \hline
        Fold 1 & 0.994597 & 0.994206 & 0.994310 & 0.994628 \\
        Fold 2 & 1.000000 & 1.000000 & 1.000000 & 1.000000 \\
        Fold 3 & 0.999151 & 0.999104 & 0.999116 & 0.999105 \\
        Fold 4 & 1.000000 & 1.000000 & 1.000000 & 1.000000 \\
        Fold 5 & 1.000000 & 1.000000 & 1.000000 & 1.000000 \\
        \hline
    \end{tabular}
\end{table}

\textbf{Test Results:} Table~\ref{tab:mobilev3_aasl_test} shows the test results for MobileNetV3, with consistently high performance across all folds.

\begin{table}[h]
    \centering
    \caption{Test Results for MobileNetV3 (AASL Dataset)}
    \label{tab:mobilev3_aasl_test}
    \begin{tabular}{|c|c|c|c|c|}
        \hline
        \textbf{Fold} & \textbf{Precision} & \textbf{Recall} & \textbf{F1-Score} & \textbf{Accuracy} \\
        \hline
        Fold 1 & 0.961020 & 0.958902 & 0.957154 & 0.962804 \\
        Fold 2 & 0.954822 & 0.955660 & 0.953747 & 0.959943 \\
        Fold 3 & 0.968383 & 0.969083 & 0.967427 & 0.971388 \\
        Fold 4 & 0.957515 & 0.958378 & 0.955710 & 0.961373 \\
        Fold 5 & 0.960098 & 0.958393 & 0.956632 & 0.962804 \\
        \hline
    \end{tabular}
\end{table}

\begin{figure}[h]
    \centering
    \includegraphics[width=0.4\textwidth]{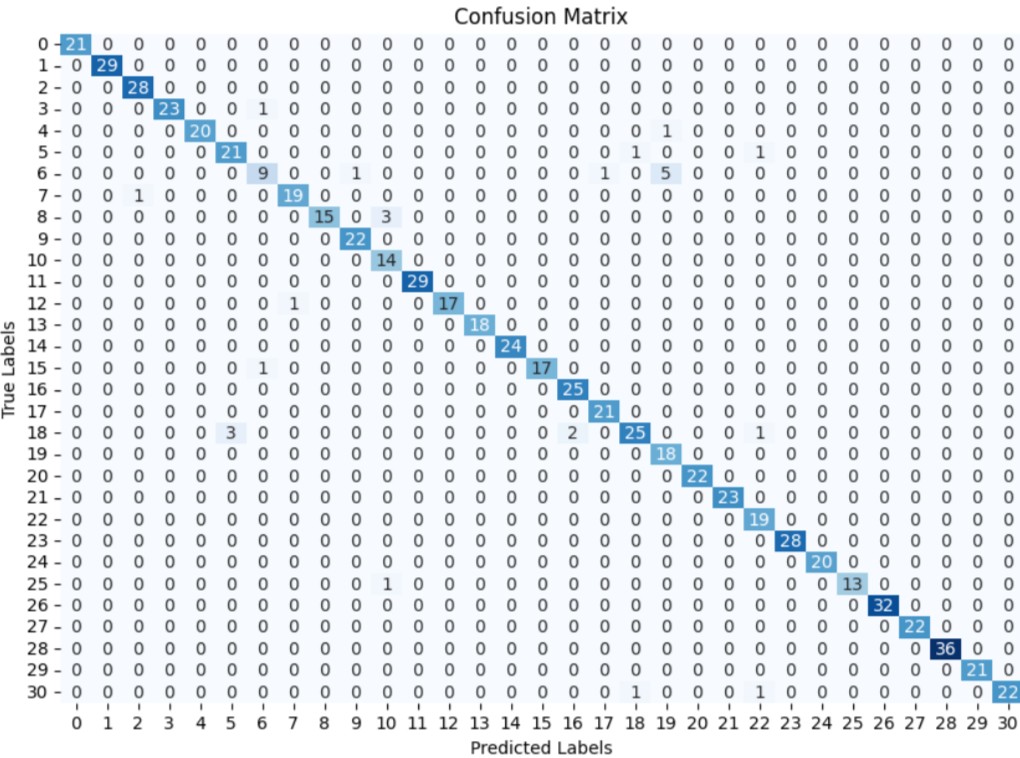}
    \caption{Confusion Matrix for MobileNetV3 (AASL Test Set, Fold 5)}
    \label{fig:cm_mobilev3_aasl}
\end{figure}

\textbf{Confusion Matrix:} The confusion matrix for MobileNetV3 (Fold 5) on the AASL test set, depicted in Figure~\ref{fig:cm_mobilev3_aasl}, shows the model's classification performance, with the majority of misclassifications occurring in similar sign gestures.

\begin{figure}[h]
    \centering
    \includegraphics[width=0.4\textwidth]{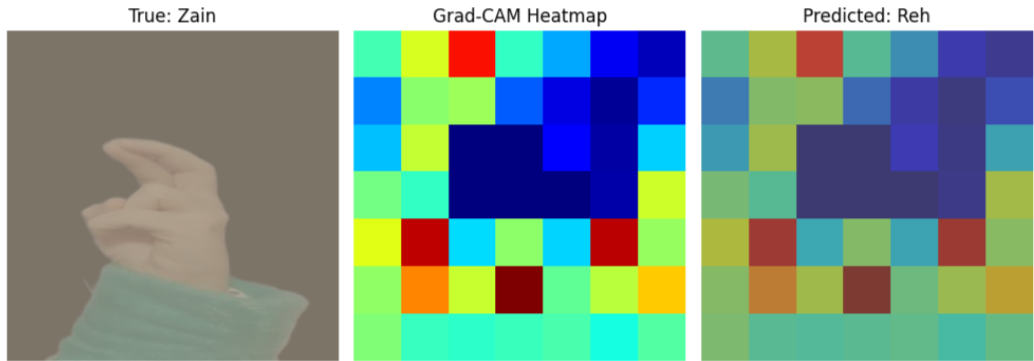}
    \caption{Grad-CAM for MobileNetV3 (AASL Test Set, Fold 5)}
    \label{fig:gradcam_mobilev3_aasl}
\end{figure}

\textbf{Grad-CAM Analysis:} Grad-CAM visualization for MobileNetV3, shown in Figure~\ref{fig:gradcam_mobilev3_aasl}, indicates that the model focuses on hand shapes and finger positions, demonstrating effective feature recognition for sign language gestures.

%%%%%%%%%%%%%%%%%%%%%%%%%%%%%%%%%%%%%%%%
\newpage

\section{ResNet50 Results}
This section presents the experimental results of the EfficientNet-B2 model on the ArSL2018 and AASL datasets. The results are divided into subsections for each dataset. For each model, we provide validation and test results, followed by analysis using confusion matrices and Grad-CAM to interpret the models' decision-making.

\subsection{ArSL2018 Dataset}

In this subsection, we present the results of our experiments on the ArSL2018 dataset using ResNet50 model.
%\subsubsection{ResNet50}
\textbf{Validation Results:} The validation results for ResNet50 are detailed in Table~\ref{tab:resnet50_arsl2018_val}, with Fold 5 achieving the highest F1-Score.

\begin{table}[h]
    \centering
    \caption{Validation Results for ResNet50 (ArSL2018 Dataset)}
    \label{tab:resnet50_arsl2018_val}
    \begin{tabular}{|c|c|c|c|c|}
        \hline
        \textbf{Fold} & \textbf{Precision} & \textbf{Recall} & \textbf{F1-Score} & \textbf{Accuracy} \\
        \hline
        Fold 1 & 0.994310 & 0.993700 & 0.993871 & 0.993906 \\
        Fold 2 & 0.997829 & 0.997748 & 0.997749 & 0.997813 \\
        Fold 3 & 0.997330 & 0.997307 & 0.997289 & 0.997344 \\
        Fold 4 & 0.997528 & 0.997517 & 0.997473 & 0.997500 \\
        Fold 5 & 0.998274 & 0.998274 & 0.998242 & 0.998281 \\
        \hline
    \end{tabular}
\end{table}

\textbf{Test Results:} The test results for ResNet50 are summarized in Table~\ref{tab:resnet50_arsl2018_test}, with the highest F1-Score observed in Fold 5.

\begin{table}[h]
    \centering
    \caption{Test Results for ResNet50 (ArSL2018 Dataset)}
    \label{tab:resnet50_arsl2018_test}
    \begin{tabular}{|c|c|c|c|c|}
        \hline
        \textbf{Fold} & \textbf{Precision} & \textbf{Recall} & \textbf{F1-Score} & \textbf{Accuracy} \\
        \hline
        Fold 1 & 0.990201 & 0.989961 & 0.989985 & 0.989750 \\
        Fold 2 & 0.991934 & 0.991998 & 0.991934 & 0.992000 \\
        Fold 3 & 0.992507 & 0.992644 & 0.992499 & 0.992500 \\
        Fold 4 & 0.992035 & 0.991651 & 0.991778 & 0.991750 \\
        Fold 5 & 0.994560 & 0.994584 & 0.994548 & 0.994500 \\
        \hline
    \end{tabular}
\end{table}

\begin{figure}[h]
    \centering
    \includegraphics[width=0.4\textwidth]{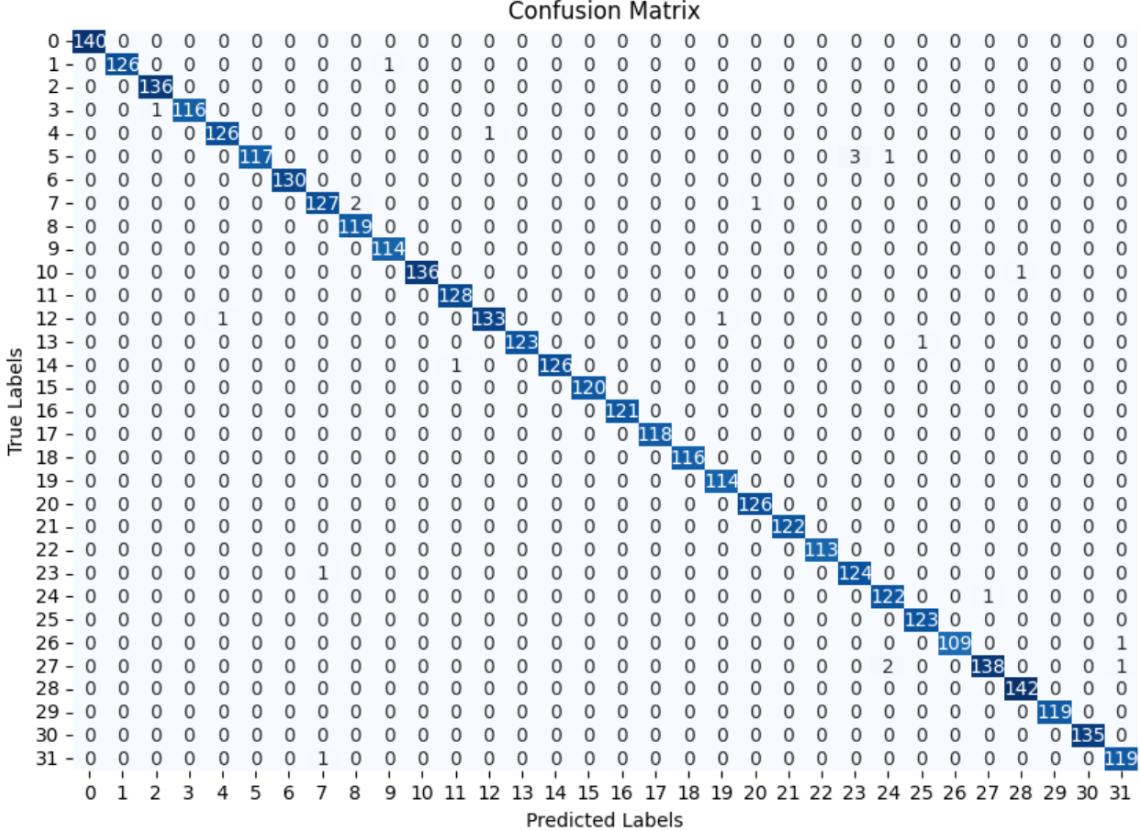}
    \caption{Confusion Matrix for ResNet50 (ArSL2018 Test Set, Fold 5)}
    \label{fig:cm_resnet50_arsl2018}
\end{figure}

\textbf{Confusion Matrix:} Figure~\ref{fig:cm_resnet50_arsl2018} displays the confusion matrix for ResNet50 (Fold 5), indicating accurate classification across most classes with only a few misclassifications.

\begin{figure}[h]
    \centering
    \includegraphics[width=0.4\textwidth]{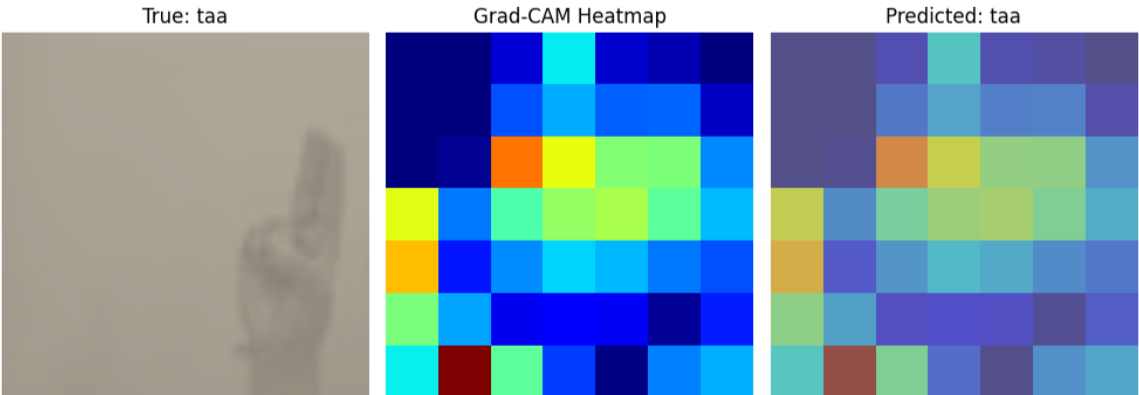}
    \caption{Grad-CAM for ResNet50 (ArSL2018 Test Set, Fold 5)}
    \label{fig:gradcam_resnet50_arsl2018}
\end{figure}

\textbf{Grad-CAM Analysis:} The Grad-CAM visualization for ResNet50 (shown in Figure~\ref{fig:gradcam_resnet50_arsl2018}) reveals that the model effectively focuses on key hand regions, providing transparent explanations for its predictions.

\subsection{AASL Dataset}

In this subsection, we present the results of our experiments on the AASL dataset using ResNet50 model.
%\subsubsection{ResNet50}
\textbf{Validation Results:} The validation results for ResNet50 on the AASL dataset are shown in Table~\ref{tab:resnet50_aasl_val}, with perfect scores achieved in Folds 1, 4, and 5.

\begin{table}[h]
    \centering
    \caption{Validation Results for ResNet50 (AASL Dataset)}
    \label{tab:resnet50_aasl_val}
    \begin{tabular}{|c|c|c|c|c|}
        \hline
        \textbf{Fold} & \textbf{Precision} & \textbf{Recall} & \textbf{F1-Score} & \textbf{Accuracy} \\
        \hline
        Fold 1 & 1.000000 & 1.000000 & 1.000000 & 1.000000 \\
        Fold 2 & 0.999128 & 0.998992 & 0.999046 & 0.999105 \\
        Fold 3 & 0.992841 & 0.992996 & 0.992738 & 0.992838 \\
        Fold 4 & 1.000000 & 1.000000 & 1.000000 & 1.000000 \\
        Fold 5 & 1.000000 & 1.000000 & 1.000000 & 1.000000 \\
        \hline
    \end{tabular}
\end{table}

\textbf{Test Results:} Table~\ref{tab:resnet50_aasl_test} shows the test results for ResNet50 on the AASL dataset, with Fold 2 achieving the best F1-Score.

\begin{table}[h]
    \centering
    \caption{Test Results for ResNet50 (AASL Dataset)}
    \label{tab:resnet50_aasl_test}
    \begin{tabular}{|c|c|c|c|c|}
        \hline
        \textbf{Fold} & \textbf{Precision} & \textbf{Recall} & \textbf{F1-Score} & \textbf{Accuracy} \\
        \hline
        Fold 1 & 0.980956 & 0.978929 & 0.979383 & 0.979971 \\
        Fold 2 & 0.986354 & 0.983633 & 0.984472 & 0.985694 \\
        Fold 3 & 0.967531 & 0.970286 & 0.967392 & 0.968526 \\
        Fold 4 & 0.982157 & 0.982460 & 0.981855 & 0.982833 \\
        Fold 5 & 0.977075 & 0.978299 & 0.976852 & 0.978541 \\
        \hline
    \end{tabular}
\end{table}

\begin{figure}[h]
    \centering
    \includegraphics[width=0.4\textwidth]{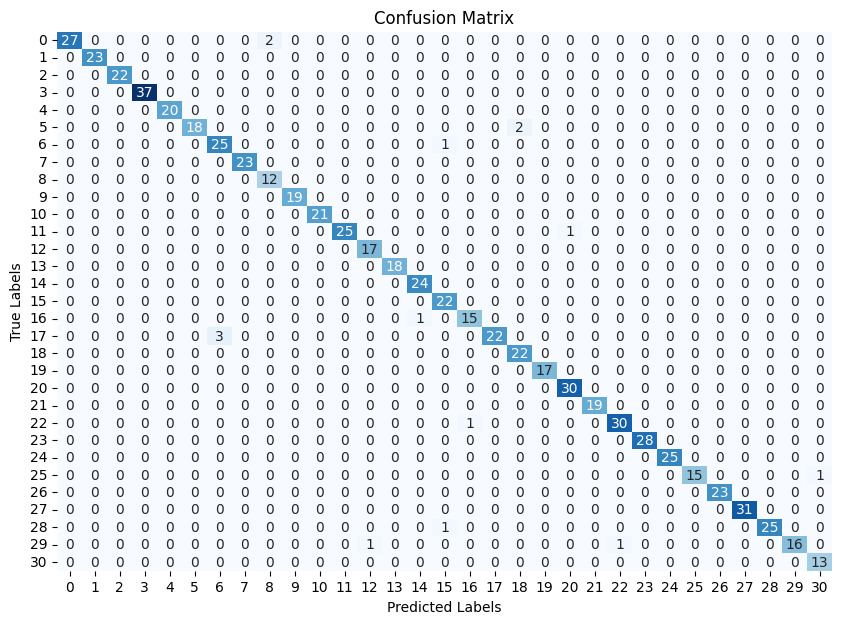}
    \caption{Confusion Matrix for ResNet50 (AASL Test Set, Fold 2)}
    \label{fig:cm_resnet50_aasl}
\end{figure}

\textbf{Confusion Matrix:} The confusion matrix for ResNet50 (Fold 2), shown in Figure~\ref{fig:cm_resnet50_aasl}, reveals a strong ability to accurately classify sign gestures, with fewer misclassifications compared to MobileNetV3.

\begin{figure}[h]
    \centering
    \includegraphics[width=0.4\textwidth]{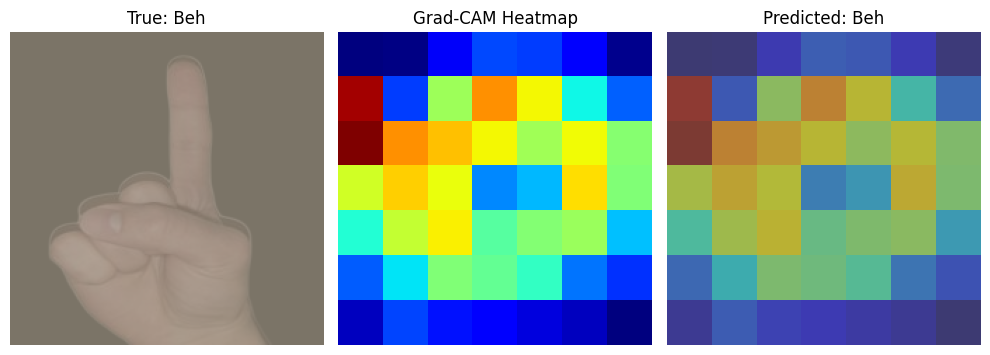}
    \caption{Grad-CAM for ResNet50 (AASL Test Set, Fold 2)}
    \label{fig:gradcam_resnet50_aasl}
\end{figure}

\textbf{Grad-CAM Analysis:} As shown in Figure~\ref{fig:gradcam_resnet50_aasl}, the Grad-CAM visualization for ResNet50 highlights critical areas of hand movements, demonstrating reliable model focus on relevant features.

\section{EfficientNet-B2  Results}
This section presents the experimental results of the EfficientNet-B2  model on the ArSL2018 and AASL datasets. The results are divided into subsections for each dataset. For each model, we provide validation and test results, followed by analysis using confusion matrices and Grad-CAM to interpret the models' decision-making.

\subsection{ArSL2018 Dataset}
In this subsection, we present the results of our experiments on the ArSL2018 dataset using EfficientNet-B2 model.
%\subsubsection{EfficientNet-B2}
\textbf{Validation Results:} The validation results for EfficientNet-B2 are shown in Table~\ref{tab:effb2_arsl2018_val}, with Fold 4 achieving the highest scores.

\begin{table}[h]
    \centering
    \caption{Validation Results for EfficientNet-B2 (ArSL2018 Dataset)}
    \label{tab:effb2_arsl2018_val}
    \begin{tabular}{|c|c|c|c|c|}
        \hline
        \textbf{Fold} & \textbf{Precision} & \textbf{Recall} & \textbf{F1-Score} & \textbf{Accuracy} \\
        \hline
        Fold 1 & 0.998147 & 0.998118 & 0.998130 & 0.998281 \\
        Fold 2 & 0.998048 & 0.997958 & 0.997990 & 0.998125 \\
        Fold 3 & 0.998090 & 0.998208 & 0.998101 & 0.998281 \\
        Fold 4 & 0.998594 & 0.998594 & 0.998594 & 0.998750 \\
        Fold 5 & 0.998455 & 0.998446 & 0.998450 & 0.998594 \\
        \hline
    \end{tabular}
\end{table}

\textbf{Test Results:} Table~\ref{tab:effb2_arsl2018_test} provides the test results for EfficientNet-B2, with consistent performance across all folds.

\begin{table}[h]
    \centering
    \caption{Test Results for EfficientNet-B2 (ArSL2018 Dataset)}
    \label{tab:effb2_arsl2018_test}
    \begin{tabular}{|c|c|c|c|c|}
        \hline
        \textbf{Fold} & \textbf{Precision} & \textbf{Recall} & \textbf{F1-Score} & \textbf{Accuracy} \\
        \hline
        Fold 1 & 0.991716 & 0.991837 & 0.991736 & 0.991750 \\
        Fold 2 & 0.992999 & 0.993450 & 0.993161 & 0.993250 \\
        Fold 3 & 0.993835 & 0.993855 & 0.993826 & 0.993750 \\
        Fold 4 & 0.994227 & 0.994498 & 0.994330 & 0.994250 \\
        Fold 5 & 0.993479 & 0.993593 & 0.993500 & 0.993500 \\
        \hline
    \end{tabular}
\end{table}

\begin{figure}[h]
    \centering
    \includegraphics[width=0.4\textwidth]{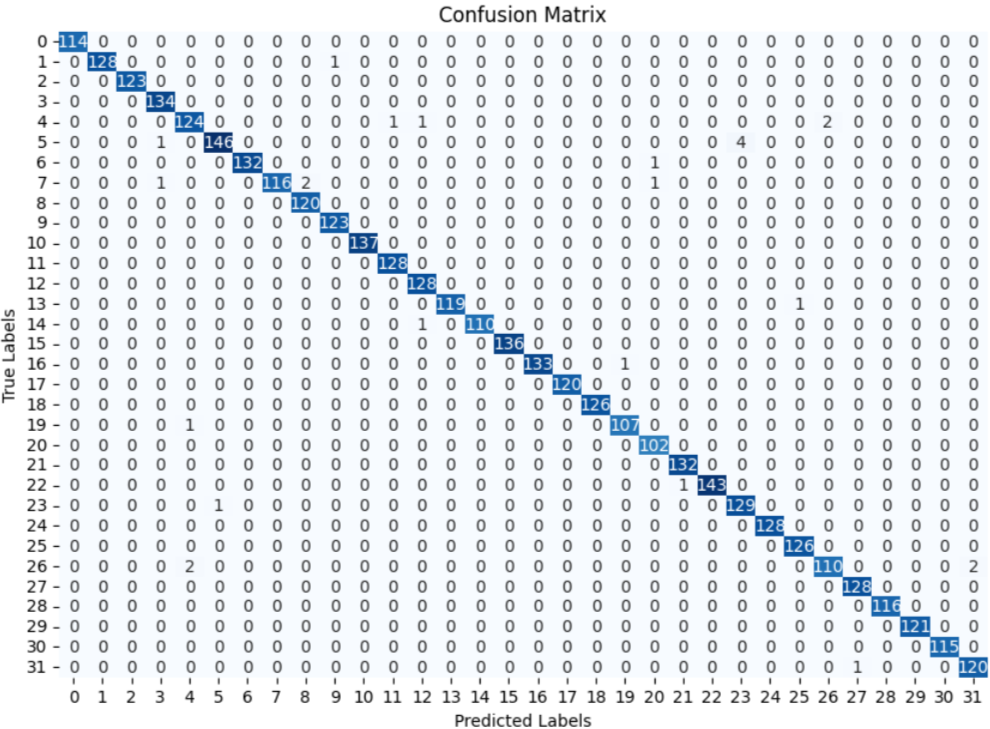}
    \caption{Confusion Matrix for EfficientNet-B2 (ArSL2018 Test Set, Fold 5)}
    \label{fig:cm_effb2_arsl2018}
\end{figure}

\textbf{Confusion Matrix:} Figure~\ref{fig:cm_effb2_arsl2018} shows the confusion matrix for EfficientNet-B2 (Fold 5), indicating accurate recognition with minimal errors.

\begin{figure}[h]
    \centering
    \includegraphics[width=0.4\textwidth]{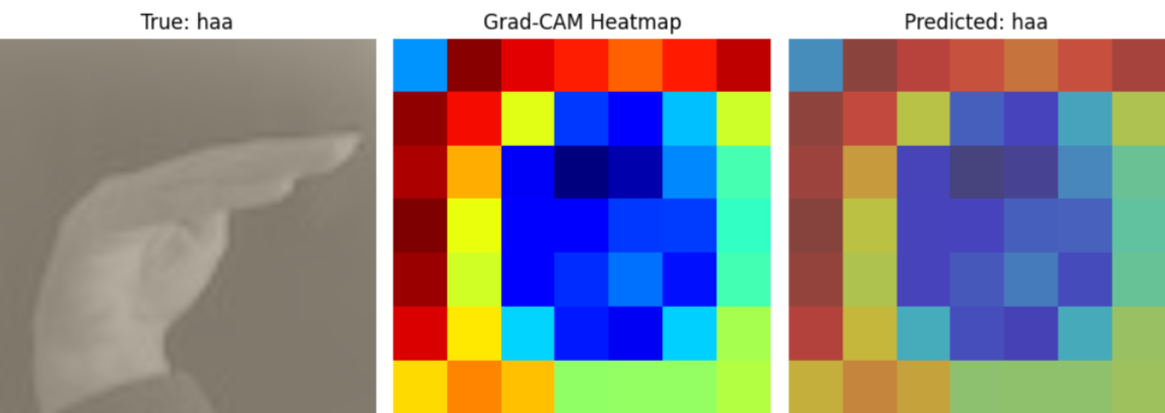}
    \caption{Grad-CAM for EfficientNet-B2 (ArSL2018 Test Set, Fold 5)}
    \label{fig:gradcam_effb2_arsl2018}
\end{figure}

\textbf{Grad-CAM Analysis:} As seen in Figure~\ref{fig:gradcam_effb2_arsl2018}, the Grad-CAM visualization for EfficientNet-B2 highlights the critical hand areas used for classification, demonstrating effective interpretability.

\subsection{AASL Dataset}

In this subsection, we present the results of our experiments on the AASL dataset using EfficientNet-B2 model.
%\subsection{EfficientNet-B2}
\textbf{Validation Results:} The validation results for EfficientNet-B2 on the AASL dataset are presented in Table~\ref{tab:effb2_aasl_val}, demonstrating perfect performance in Folds 3, 4, and 5.

\begin{table}[h]
    \centering
    \caption{Validation Results for EfficientNet-B2 (AASL Dataset)}
    \label{tab:effb2_aasl_val}
    \begin{tabular}{|c|c|c|c|c|}
        \hline
        \textbf{Fold} & \textbf{Precision} & \textbf{Recall} & \textbf{F1-Score} & \textbf{Accuracy} \\
        \hline
        Fold 1 & 0.999213 & 0.999403 & 0.999300 & 0.999105 \\
        Fold 2 & 0.999078 & 0.998710 & 0.998874 & 0.999105 \\
        Fold 3 & 1.000000 & 1.000000 & 1.000000 & 1.000000 \\
        Fold 4 & 1.000000 & 1.000000 & 1.000000 & 1.000000 \\
        Fold 5 & 1.000000 & 1.000000 & 1.000000 & 1.000000 \\
        \hline
    \end{tabular}
\end{table}

\textbf{Test Results:} The test results for EfficientNet-B2, shown in Table~\ref{tab:effb2_aasl_test}, reveal strong performance across all folds, with Fold 4 achieving the highest F1-Score.

\begin{table}[h]
    \centering
    \caption{Test Results for EfficientNet-B2 (AASL Dataset)}
    \label{tab:effb2_aasl_test}
    \begin{tabular}{|c|c|c|c|c|}
        \hline
        \textbf{Fold} & \textbf{Precision} & \textbf{Recall} & \textbf{F1-Score} & \textbf{Accuracy} \\
        \hline
        Fold 1 & 0.983857 & 0.980665 & 0.981593 & 0.982833 \\
        Fold 2 & 0.987657 & 0.988463 & 0.987757 & 0.988555 \\
        Fold 3 & 0.987088 & 0.988084 & 0.987294 & 0.988555 \\
        Fold 4 & 0.989727 & 0.988299 & 0.988538 & 0.988555 \\
        Fold 5 & 0.989385 & 0.989718 & 0.989332 & 0.989986 \\
        \hline
    \end{tabular}
\end{table}

\begin{figure}[h]
    \centering
    \includegraphics[width=0.4\textwidth]{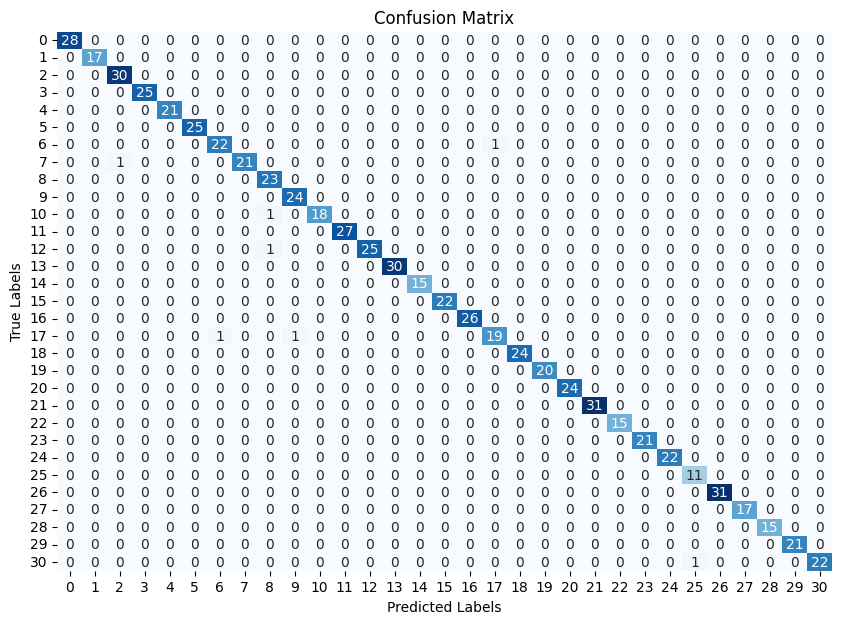}
    \caption{Confusion Matrix for EfficientNet-B2 (AASL Test Set, Fold 4)}
    \label{fig:cm_effb2_aasl}
\end{figure}

\textbf{Confusion Matrix:} The confusion matrix for EfficientNet-B2 (Fold 4) is depicted in Figure~\ref{fig:cm_effb2_aasl}, showing minimal misclassifications and strong classification accuracy.

\begin{figure}[h]
    \centering
    \includegraphics[width=0.4\textwidth]{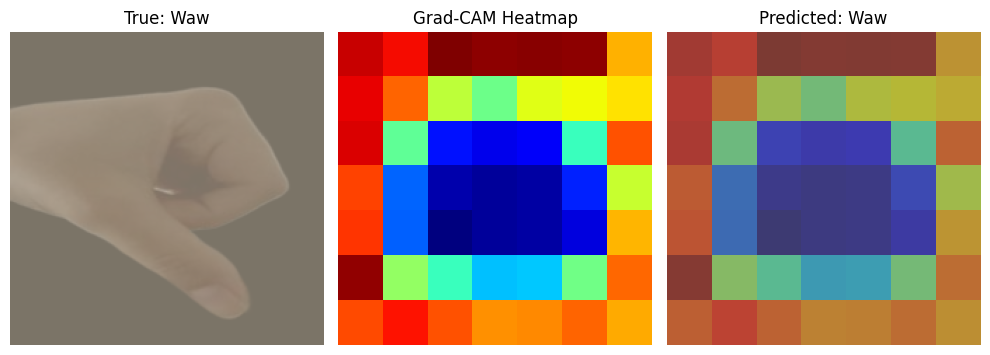}
    \caption{Grad-CAM for EfficientNet-B2 (AASL Test Set, Fold 4)}
    \label{fig:gradcam_effb2_aasl}
\end{figure}

\textbf{Grad-CAM Analysis:} The Grad-CAM visualization for EfficientNet-B2, presented in Figure~\ref{fig:gradcam_effb2_aasl}, highlights the model's focus on the critical parts of hand gestures, confirming its interpretability.

\section{Comparison with State-of-the-Art Methods} \label{sec:comparison}
Our approach outperforms several existing models in Arabic Sign Language (ArSL) recognition, as outlined in Table~\ref{tab:comparison_with_others}. On the ArSL2018 dataset, our EfficientNet-B2 model achieves a test accuracy of 99.48\%, surpassing previous models like those by Hu et al. \cite{hu2022sign}, Abdelghfar et al. \cite{abdelghfar2023model}, and Al Nabih et al. \cite{alnabih2024arabic}. Similarly, on the AASL dataset, EfficientNet-B2 reaches a test accuracy of 98.99\%, improving upon results from El Baz et al. \cite{el2024deep} and others.

\begin{table}[ht]
\centering
\caption{Comparison with Other Studies on ArSL2018 and AASL Datasets}
\label{tab:comparison_with_others}
\begin{tabular}{|p{2.5cm}|p{1.5cm}|p{1.5cm}|}
\hline
\textbf{Study} & \textbf{Dataset} & \textbf{Test Accuracy (\%)} \\
\hline
Hu et al. \cite{hu2022sign} (EfficientNetB4) & ArSL2018 & 95.00 \\
Abdelghfar et al. \cite{abdelghfar2023model} (QSLRS-CNN) & ArSL2018 & 97.31 \\
Al Nabih et al. \cite{alnabih2024arabic} (Vision Transformers) & ArSL2018 & 99.30 \\
Lahiani et al. \cite{lahiani2024exploring} (MobileNetV2) & ArSL2018 & 96.00 \\
Hassan et al. \cite{hassan2024detection} (PCA, LDA, KNN) & ArSL2018 & 86.40 \\
\textbf{Our Approach (EfficientNet-B2)} & ArSL2018 & \textbf{99.48} \\
\hline
El Baz et al. \cite{el2024deep} (CNN) & AASL & 97.40 \\
Renjith et al. \cite{renjith2024sign} (Spatio-temporal approach) & ArSL & 89.46 \\
\textbf{Our Approach (EfficientNet-B2)} & AASL & \textbf{98.99} \\
\hline
\end{tabular}
\end{table}

These improvements in our models can be attributed to:
\begin{itemize}
    \item \textbf{Advanced Data Augmentation:} Our preprocessing pipeline includes extensive data augmentation, which improves model generalization and addresses class imbalances.
    \item \textbf{Explainable AI (XAI):} The use of Grad-CAM visualizations enhances interpretability, making it easier to understand model decisions.
    \item \textbf{Cutting-Edge Deep Learning Architectures:} EfficientNet-B2 captures intricate hand gestures more effectively, resulting in higher accuracy.
\end{itemize}

These results demonstrate the robustness and scalability of our approach in recognizing Arabic Sign Language, making it suitable for broader applications, such as healthcare and education, where transparency and reliability are crucial. The use of explainable AI techniques ensures the models are interpretable and can be adapted to other sign languages as well.

\section{Conclusion and Future Work}\label{sec:conclusion}

This study successfully developed an advanced system for Arabic Sign Language (ArSL) recognition using cutting-edge deep learning models, including MobileNetV3, ResNet50, and EfficientNet-B2, combined with Explainable AI (XAI) techniques like Grad-CAM to enhance transparency and interpretability. The proposed system demonstrated superior accuracy and F1-score compared to existing approaches on the ArSL2018 and AASL datasets, with EfficientNet-B2 achieving the highest performance. Robust data preprocessing, extensive data augmentation, and stratified 5-fold cross-validation contributed to balanced learning across diverse samples, making the model suitable for deployment in healthcare, education, and inclusive communication technologies.

Future work can explore the integration of the proposed model into real-time applications, such as video-based communication platforms, to enhance accessibility and inclusivity. Further research into transformer-based architectures, such as Vision Transformers (ViT) or hybrid models, could enhance performance by capturing long-range dependencies in sequential gestures. Expanding the system to a multi-modal framework that integrates audio, text, and visual signals could improve interpretability and versatility. Extending the model to recognize other sign languages—like American Sign Language (ASL) or Chinese Sign Language (CSL)—can broaden its applicability, creating a global, multilingual sign language recognition framework. Improved data collection, especially for underrepresented classes, will enhance robustness and address class imbalances. Future implementations could also focus on creating explainable user interfaces (UIs) that provide clear model interpretations for users, fostering trust in AI decisions. Additionally, integrating the model with IoT devices, such as smart glasses or handheld devices, could offer immersive, real-time communication experiences.

\bibliographystyle{IEEEtran}
\bibliography{cite}

%\tableofcontents

\end{document}